\documentclass[lettersize,journal]{IEEEtran}
\usepackage{amsmath,amsfonts}
\usepackage{algorithmic}
\usepackage{algorithm}
\usepackage{array}
\usepackage[caption=false]{subfig}
\usepackage{textcomp}
\usepackage{stfloats}
\usepackage{url}
\usepackage{verbatim}
\usepackage{graphicx}
\usepackage{hyperref}
\usepackage{bbm}
\usepackage{tikz}
\usetikzlibrary{shapes,snakes}
\usetikzlibrary{automata, arrows, positioning, calc}
\usepackage{tabularx}
\usepackage{booktabs}
\usepackage{pgfplots}
\usepackage{multirow}
\usepackage[numbers]{natbib}
\usepackage{float}

\DeclareMathOperator{\E}{\mathbb{E}}

\pgfplotsset{compat=1.16}
\begin{document}

\title{A Probabilistic Model for  Skill Acquisition with Switching Latent Feedback Controllers}

\author{Juyan Zhang, Dana Kuli\'c and Michael Burke\\
  Department of Electrical and Computer Systems Engineering \\
  Faculty of Engineering \\
  Monash University
  Australia
}

\markboth{Journal of \LaTeX\ Class Files,~Vol.~14, No.~8, August~2021}%
{Shell \MakeLowercase{\textit{et al.}}: A Sample Article Using IEEEtran.cls for IEEE Journals}


\maketitle

\begin{abstract}
Manipulation tasks often consist of subtasks, each representing a distinct skill. Mastering these skills is essential for robots, as it enhances their autonomy, efficiency, adaptability, and ability to work in their environment. Learning from demonstrations allows robots to rapidly acquire new skills without starting from scratch, with demonstrations typically sequencing skills to achieve tasks. Behaviour cloning approaches to learning from demonstration commonly rely on mixture density network output heads to predict robot actions. In this work, we first reinterpret the mixture density network as a library of feedback controllers (or skills) conditioned on latent states. This arises from the observation that a one-layer linear network is functionally equivalent to a classical feedback controller, with network weights corresponding to controller gains. We use this insight to derive a probabilistic graphical model and a new evidence lower bound that combines these elements, describing the skill acquisition process as segmentation in a latent space, where each skill policy functions as a feedback control law in this latent space. Our approach significantly improves not only the task success rate, but also robustness to observation noise when trained with human demonstrations. Our physical robot experiments further show that the induced robustness improves model deployment on robots.
\end{abstract}


\section{Introduction}

\IEEEPARstart{A} proficient robot should excel at performing diverse tasks and reusing its skills across various scenarios. A common strategy for handling tasks of varying complexity is to decompose these into simpler stages, requiring intermediate goals to be met over shorter time horizons \cite{kroemer_towards_2015}. This approach not only simplifies the learning process but also results in the development of reusable skills.

For instance, when a robot prepares coffee, the process involves several distinct skills. The robot first attempts to reach the cup, and when it and the cup are aligned at the same position, the subtask of pouring the coffee begins. This pattern continues for subsequent subtasks, such as stirring and cup carrying. This scenario reveals two key insights: first, all these subtasks require moving the robot and objects to designated target positions; second, the transitions between these subtasks are determined by the states of both the robot and the objects. Therefore, it is logical to establish a skill policy as a feedback control law that considers the states and set points of the robot and relevant objects in the environment, with transitions determined by this joint state representation. This work considers a family of feedback control laws parametrised by controller gains and goals or controller set points. Completing a complex task involves applying a sequence of feedback control laws that guide the robot and its environment towards a series of target states. We define a skill as the set of motions that can be realised by a specific feedback controller, and the composition of skills as the act of switching between different feedback controllers.

A widely used network structure for learning tasks in various robotics scenarios ranging from autonomous driving \cite{chai2020multipath, choi2019distributional, huang2022efficient} to robot manipulation \cite{bishop1994mixture, angelov2020composing, liu_libero_2023,pignat2019bayesian,wilson2020learning,rahmatizadeh2018virtual, prasad2024moveint}, is the mixture density network (MDN), a neural network that predicts a distribution over actions using a Gaussian mixture model. However, MDNs can easily overfit due to the flexibility of mixture models \cite{cui2019multimodal, makansi2019overcoming}.  In this work, we first reinterpret a one-layer linear network as a classical feedback controller, acting on a hidden or latent state. This in turn allows us to reinterpret the MDN as a switching library of skill policies conditioned on a shared latent state. We use this interpretation to derive a probabilistic graphical model describing the skill acquisition process as a segmentation of a latent space, where each skill policy is a feedback control law in the corresponding space. Our hypothesis is that enforcing latent feedback structures into a model helps bound the model to states seen at training, resulting in more robust performance.

Our model framework\footnote{Project Page: \url{https://sites.google.com/view/skillacquisition/home}}  builds on Stochastic neural networks \cite{florensa2017stochastic} and Switching density networks \cite{burke_hybrid_2020}, and improves them in two ways: 
Firstly, our model uses latent states and therefore allows for different types of measurements or even modalities of input instead of explicitly defined states like positions or orientations. Secondly, it provides complete probabilistic derivations that result in policies that allow for more robust control when dealing with noisy state measurements. The key contributions of this work are summarized as follows:
\begin{itemize}
    \item We provide a way to interpret a one-layer neural network as a classical feedback controller in latent space.
    \item We provide a probabilistic approach to learning a sequence of feedback laws in the latent space with a new evidence lower bound for training.
    \item We evaluate the proposed approach and compare it to behaviour cloning and Mixture Density Networks baselines in both simulations and physical experiments, and show that our model improves task success rate, robustness to noise, and the skill transitions and interpretability of the learned skills in tasks specified by natural language descriptions.
\end{itemize}

\begin{figure*}[!htb]
    \centering
    \subfloat[Model structure]{
        \includegraphics[trim=4cm 7.5cm 4cm 2cm, clip, scale=0.45]{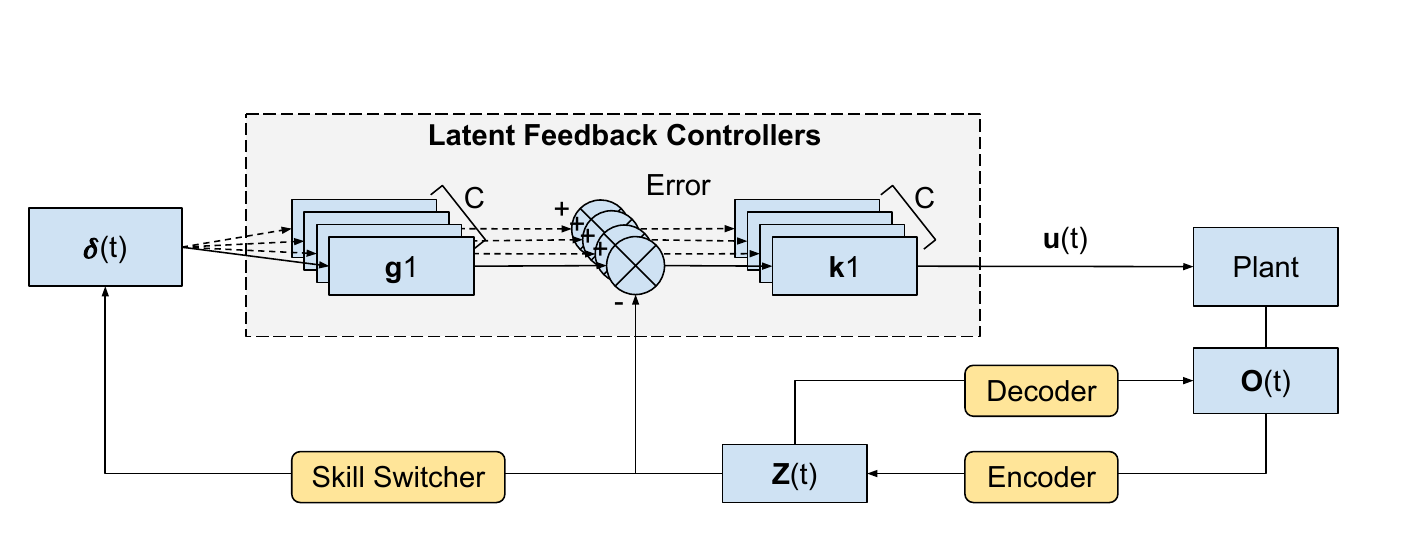}
        \label{fig: Model Structure}
    }%
    \hfil
    \subfloat[Probabilistic Model]{
       \begin{tikzpicture}[node distance=1.4cm, baseline={(0,-2.5)}]
        \tikzstyle{block} = [circle, draw, fill=blue!20, 
            text width=1em, text centered, rounded corners, minimum height=1em]
        \tikzstyle{fixed} = [circle, draw, fill=gray!20, 
            text width=1em, text centered, rounded corners, minimum height=1em]
        \tikzstyle{latent} = [circle, draw, fill=yellow!50, 
            text width=1em, text centered, rounded corners, minimum height=1em]
        \tikzstyle{latent_wide} = [rectangle, draw, fill=gray!50, 
            text width=3em, text centered, rounded corners, minimum height=1em]

        \node [latent_wide] (goal_gain) {$\boldsymbol{g}_t$, $\boldsymbol{k}_t$};
        \node [latent, left of=goal_gain] (index) {$\boldsymbol{\delta}_t$};
        \node [latent, below=of $(goal_gain)!0.5!(index)$] (state) {$\boldsymbol{z}_t$};
        \node [block, right of=goal_gain] (action) {$\boldsymbol{u}_t$};
        \node [block, below=of $(goal_gain)!0.5!(action)$] (obs) {$\boldsymbol{o}_t$};

        \draw[->] (index) -- (goal_gain);
        \draw[->] (state) -- (index);
        \draw[->] (state) -- (action);
        \draw[->] (goal_gain) -- (action);
        \draw[->] (state) -- (obs);

        \draw[->, dashed, red] (obs) to [out=-135, in=-45] (state);
        \draw[->, dashed, red] (action) to [out=135, in=45] (index);
    \end{tikzpicture}
    \label{fig: generation process}
    }%
    \caption{(a) Model Structure Overview:  We model skills as a set of feedback controllers in latent space. The control signals are predicted using a latent controller given a latent gain, a latent goal and the current latent state. Yellow modules are neural networks, while others denote components in the feedback loop.  $\boldsymbol{o}(t)$ is observation at time $t$. The encoder neural network maps $\boldsymbol{o}(t)$ to the latent space $\boldsymbol{z}(t)$. The skill switcher network predicts the skill index that switches on the $\boldsymbol{\delta}$th latent feedback controller with reference point $\boldsymbol{g}[\boldsymbol{\delta}]$ and the gain $\boldsymbol{k}[\boldsymbol{\delta}]$. The robot receives the control signal $\boldsymbol{u}$ for execution; (b) Probabilistic graphical model: Solid arrows indicate the generation process. Red dashed arrows indicate the approximating process. $\boldsymbol{g}_t, \boldsymbol{k}_t$ are not random variables. The blue circles are the observed random variables and the yellow circles are the inferred latent random variables.}
\end{figure*}

\section{Related Work}

As highlighted by \cite{kroemer_review_2021}, efficient skill acquisition depends on the autonomous discovery of the underlying skill hierarchy. To achieve this, it is crucial to not only learn skill segmentation (when and where to apply skills) but also to learn the appropriate control policies that underlie each skill simultaneously. 

\paragraph{What to segment} 
Numerous approaches have been proposed to segment skills. One line of work is based on similarity. These methods identify skills by examining similarities in different spaces, such as the state space \cite{tanneberg_skid_2021}, policy parameter space \cite{daniel_probabilistic_2016}, reward function space \cite{krishnan2019swirl, ranchod2015nonparametric}, and more recently, task space through language embeddings \cite{liang_skilldiffuser_2024}. Similarity-based approaches have also been applied in hierarchical and options-based reinforcement learning \cite{sutton1999between} \cite{stolle2002learning}. While these approaches provide valuable insights, they often struggle with complex tasks. Similarity measures can fail to generalize across diverse states or reward structures, and reinforcement learning methods frequently face challenges related to sample efficiency and computational complexity.

Another line of research models preconditions for skill switching with change point detection \cite{fitzpatrick2006reinforcing}, which segments skills by identifying shifts in state dynamics. While effective for simple proprioceptive states such as positions and velocities, this method is less suited for high-dimensional data like images. Moreover, trajectory-segmentation-based methods \cite{tanneberg_skid_2021} can fail on geodesic motions, while event-based approaches are prone to errors if event detection is subtle. Both types are also sensitive to sensor noise, further limiting their robustness.

A common issue with both similarity-based and change point detection methods is their sensitivity to noise and inefficiency when handling complex, multimodal data. To address these limitations, we propose a new approach that examines skill similarity based on the causal relationships between states and actions, deriving posterior distributions over skills through a probabilistic framework. By incorporating latent states, our method effectively integrates multimodal information, resulting in more robust skill segmentation.

\paragraph{How to segment} Mixture models  \cite{bishop1994mixture} have been commonly used to learn segmentation of skills. Among those, the mixture density network \cite{bishop1994mixture} (MDN) is one of the most commonly used models as the action head (the final layer of a neural network policy) due to its ability to model multimodal output distributions. In the context of skill acquisition with a predefined number of skills $C$, the MDN learns to switch between different control signal $\boldsymbol{u}_t$ modes according to observations or states $\boldsymbol{z_t}$, 
\begin{align}
    q(\boldsymbol{u}_t|\boldsymbol{z}_t) = \sum_{i=1}^C \pi_i(\boldsymbol{z}_t) \mathcal{N}(\boldsymbol{u}_t|\mu_i(\boldsymbol{z}_t), \Sigma_i(\boldsymbol{z}_t)) \label{eq: mdn}
\end{align}
where the mean $\mu_i(\boldsymbol{z}_t)$, variance $\Sigma_i(\boldsymbol{z}_t)$ and the normalised weights $\pi_i(\boldsymbol{z}_t)$ are predicted using neural networks. The MDN has been extended to include proportional control law structures. These switching density networks  \cite{burke_hybrid_2020} identify and learn a sequence of goals and gains in the state spaces and use a set of independent PID control laws to generate control signals for each action dimension. Unfortunately, this approach applies feedback based on the true state of a robot's end effectors which limits applications and does not scale to the full multivariable control problem. The Newtonian Variational AutoEncoders  \cite{jaques_newtonianvae_2021} attempted to remedy this using a two-stage approach. This model first learned a locally linear latent dynamics model from a large pool of data collected by motor babbling.  A second stage then inferred goals on this latent manifold along with proportional controller gains from a set of demonstrated data. Our approach is similar, but does not require a dynamics model or two stages of learning, and extends the proportional control laws to a multivariable feedback control law. This is accomplished by a new derivation of an evidence-lower bound that specifically considers a switching feedback control law structure.  

{\color{red} \paragraph{Variational Autoencoders(VAE)} Our derivation relies on variational Autoencoding  \cite{kingma2013auto}, an approach to learning latent variables using variational optimization, by maximising an Evidence Lower Bound (ELBO). Variational autoencoders have previously been combined with Gaussian mixture models to learn richer latent representations \cite{dilokthanakul2016deep}. However, these approaches typically rely on a categorical prior over mixture components and do not incorporate the \textbf{feedback control structure} considered in our work. In contrast, our model introduces a \textbf{switching mechanism} that explicitly integrates the feedback mechanism into the latent space, enabling the selection of control policies based on latent state observations.}

\section{Methodology}
\subsection{Problem Formulation}
Given $N$ demonstration samples consisting of observation $\boldsymbol{o}_{1:N, 1:T}$ and control signal $\boldsymbol{u}_{1:N, 1:T}$ pairs, we hypothesise that the demonstrations are generated by a sequence of skills, represented as controllers in a latent space.  We want to identify which skill is active at time step $t$, represented as the skill index $\boldsymbol{\delta}_{it}$, and the corresponding skill policies $\pi_{\boldsymbol{\delta}_{it}}(\boldsymbol{z}_t)$, where $\boldsymbol{z}_t$ is the state at time $t$. We propose to learn these skills and latent state representation in a generative way, by maximizing the joint probability of the demonstration sequences, $P(\boldsymbol{o}_{1:N, 1:T}, \boldsymbol{u}_{1:N, 1:T})$.

In the above, $\boldsymbol{o}_{it}$ denotes the observation at time $t$ for the $i^{th}$ demonstration, $\boldsymbol{u}_{it}$ the control signal generated by the policy at time $t$ in the $i^{th}$ demonstration and which, when applied to the robot,  will result in a new observation $\boldsymbol{o}_{i,t+1}$. $T$ denotes the maximum demonstration sequence length and $N$ the number of demonstrations. For the $i$th demonstration trajectory, $\pi_{\delta_{t}}$ is the $\boldsymbol{\delta}_{t}$th skill policy. In reality, we usually don't have direct access to the state $\boldsymbol{z}_t$, so it is estimated from $\boldsymbol{o}_{t}$. For the rest of the paper, we will ignore the trajectory index $i$ to simplify the notation. 

As an example, consider a robot performing a table-top manipulation task, such as assembling a small toy. The robot receives visual observations from a camera and control signals that guide its actions. The observations $\boldsymbol{o}_t$ could be images of the workspace, and the control signals $\boldsymbol{u}_t$ could be the robot's joint or end-effector velocities. In this scenario, we want the robot to learn how to segment its actions into meaningful skills, like picking up a piece, placing it in the correct position, and tightening a screw. Each segment of the task can be represented by a skill index $\boldsymbol{\delta}_{t}$ and each skill has a corresponding policy $\pi_{\boldsymbol{\delta}_{t}}(\boldsymbol{z}_t)$ that dictates the robot's actions based on its current state $\boldsymbol{z}_t$ and $\boldsymbol{\delta}_{t}$, which subsequently determines the goal $\boldsymbol{g}_t$ and gain $\boldsymbol{k}_t$.

\subsection{Proposed Approach}
We first introduce the feedback control law interpretation of fully connected neural networks and show that the network can be viewed as a feedback controller in a latent space. Assuming that the latent state $\boldsymbol{z}$ is a Markovian state or a representation of the entire state of the environment, we introduce a model that enforces this assumption and extends to a policy that switches between multiple linear feedback control laws. Finally, we formulate a probabilistic graphical model to integrate the switching modules and feedback control and train it using variational inference. After the model is trained, we are able to generate control signals using the learned policies.

\subsubsection{Fully Connected neural networks as feedback control laws}
We start by noting that a linear layer in a neural network can be rewritten as a form of feedback control law.
\begin{align}
    \boldsymbol{u}_t &= \boldsymbol{W}\boldsymbol{z}_t + \boldsymbol{b} \notag \\
      &= (-\boldsymbol{W})(-\boldsymbol{W}^{\dagger}\boldsymbol{b}-\boldsymbol{z}_t) \notag \\
      &= \boldsymbol{K}(\boldsymbol{g} - \boldsymbol{z}_t) \label{eq: feedback laws}
\end{align}
where $\boldsymbol{K}=-\boldsymbol{W}$ is a gain matrix, $\boldsymbol{g} = -\boldsymbol{W}^{\dagger}\boldsymbol{b}$ is analogous to a set point or goal in classical control with $\boldsymbol{W}^{\dagger}$ as the Moore–Penrose inverse of the weight matrice $\boldsymbol{W}$ and $\boldsymbol{z_t}$ is a latent state representation. 

The feedback control law interpretation above assumes that the latent state $\boldsymbol{z}_t$ is a Markovian state of the entire state of the environment. However, in standard neural network architectures, this assumption does not generally hold, as the hidden units are not explicitly constrained to represent the states. Below, we introduce a model that enforces this assumption and uses the resulting latent space to form a policy that switches between multiple feedback control laws. 

\subsubsection{Switching latent feedback control laws}
Given a latent feedback controller in the form of \autoref{eq: feedback laws} with known state, gains and control signals, an estimate of the latent goal $\boldsymbol{g}$ based on the data available at time t, $(\boldsymbol{z}_t, \boldsymbol{u}_t)$ is:
{\color{red}$$\boldsymbol{g}_t = \boldsymbol{K}^{\dagger}\boldsymbol{u}_t + \boldsymbol{z}_t$$
Here $\boldsymbol{K}^{\dagger}$ denotes the Moore–Penrose inverse of the gain matrix $\boldsymbol{K}$, which accounts for potential differences in dimensionality between the action and state spaces.}

We assume that each skill has its own distinct goal and gain. For a trajectory of length T, we have T estimates of goals. In the latent space, the trajectory samples that try to achieve the same subtask share a common subgoal. Furthermore, for multiple demonstrations of the same task, our objective is to segment the trajectories by identifying skills (and therefore goals) that can be shared across different demonstrations. The segmentation of trajectories is achieved by clustering these common latent goals. To model this, we assume that the latent goals follow a Gaussian mixture distribution, which is captured using the MDN.  
\begin{align}
    p(\boldsymbol{g}_t|\boldsymbol{z}_t) = \sum_{i=1}^C \pi_i(\boldsymbol{z}_t) \mathcal{N}(\boldsymbol{g}_t|\boldsymbol{\mu}_i(\boldsymbol{z}_t), \boldsymbol{\Sigma}_i(\boldsymbol{z}_t)) \label{eq: goal cluster}
\end{align}

Combining Equation \ref{eq: feedback laws} and Equation \ref{eq: goal cluster}, the control signals can be formulated as follows.

{\color{red}
\begin{align}
    p(\boldsymbol{u}_t|\boldsymbol{z}_t) =  \sum_{i=1}^C \pi_i(\boldsymbol{z}_t) \mathcal{N}(\boldsymbol{u}_t|\boldsymbol{K}_i(\boldsymbol{g}_t - \boldsymbol{z}_t), \boldsymbol{K}_i\boldsymbol{\Sigma}(\boldsymbol{z}_t)\boldsymbol{K}^T_i) \label{eq: 2}
\end{align}
}

As shown above, feedback control laws can be formulated as fully connected neural networks. If the new covariance matrix is predicted by a neural network $\hat{\boldsymbol{\Sigma}}_i$, without loss of generality, Equation \ref{eq: 2} can be further simplified as 
\begin{align}
    p(\boldsymbol{u}_t|\boldsymbol{z}_t) &=  \sum_{i=1}^C \pi_i(\boldsymbol{z}_t) \mathcal{N}(\boldsymbol{u}_t|\boldsymbol{\mu}_i(\boldsymbol{z}_t), \hat{\boldsymbol{\Sigma}}_i(\boldsymbol{z}_t)) \label{eq: 3}
\end{align}

Notice that $\pi_i(\boldsymbol{z}_t)$ predicts the next skill at time t without verifying if $\boldsymbol{g}_t$ has been achieved. This implies that each skill can transition to the next without necessarily reaching its goal, allowing for more flexible behaviours such as blending between skills, especially when the number of skills is limited.

Recognizing the equivalence between Equation \ref{eq: 3} above and Equation \ref{eq: mdn} for the MDN, it becomes clear that MDN naturally contains all the necessary components for our design. 

However, there are no constraints to enforce the switching feedback control law interpretation in a vanilla MDN. To address this, we introduce a probabilistic graphical model to enforce the desired design, which leads to our final model structure, shown in Figure \ref{fig: Model Structure}.

\begin{figure*}[!htb]
\centering
    \subfloat[Libero Tasks]{
    \includegraphics[width=1.2\columnwidth]{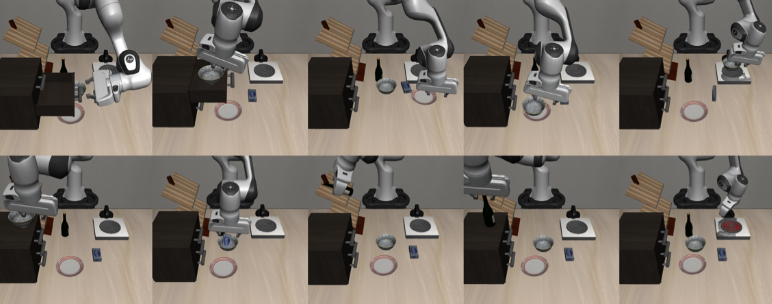}
    \label{fig: libero Dataset}
    }%
    \hfil
    \subfloat[FetchPush Tasks]{
        \includegraphics[width=0.48\columnwidth]{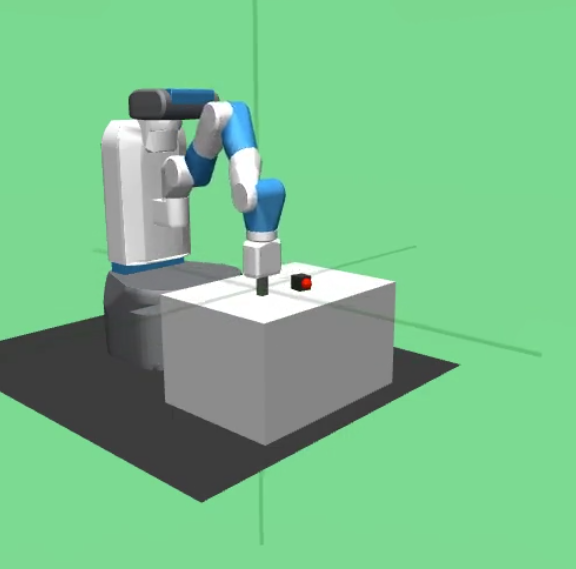}
        \label{fig: fetchpush}
    }%
    \caption{Simulated Environments for evaluations. (a) Franka Kitchen \cite{liu_libero_2023}: Multitask environment with demonstrations of manipulating different objects in the Kitchen Scene for realistic and scalable evaluations (b) FetchPush Task \cite{towers_gymnasium_2023}: The Fetch robot pushes the block to the target position for thorough performance analysis of models. }
\end{figure*}

\subsubsection{Probabilistic model of skill acquisition}

Given a predefined number $\boldsymbol{C}$ of skills to discover, we aim to segment the demonstration into skills and learn the corresponding policy for each skill. We assume that control signals for each skill are generated using a linear feedback control law in a latent space shared among skills. Sharing this latent space potentially enables smoother transitions between skills by unifying their representations. By representing multiple skills in the same latent space, we hope that the model can more easily switch or blend between them.

We formulate our skill acquisition model into the three modules illustrated in Figure \ref{fig: Model Structure}. First, an encoder encodes observations to latent states. Second, a skill transition model segments the whole trajectory into different parts and lastly, a latent linear feedback controller acts on the latent space to output control signals based on latent goal and gain matrices.

The autoencoder structure ensures that the latent representation is a latent \textit{state}, and determines which skill to execute, along with the corresponding gains and goals that generate appropriate control signals. 

\paragraph{Generation Process}
This generative process is shown in Figure \ref{fig: generation process}. We assume conditional independence of observations $\boldsymbol{o}_t$ given $\boldsymbol{z}_t$, that the conditional distribution is Gaussian and that $\boldsymbol{o}_t$, the observation at time $t$, follows a normal distribution with fixed learnable standard deviation,
\begin{align}
    p(\boldsymbol{o}_t | \boldsymbol{z}_t, \boldsymbol{\delta}_t, \boldsymbol{u}_t) &= p(\boldsymbol{o}_t|\boldsymbol{z}_t) \\
    \boldsymbol{o}_t|\boldsymbol{z}_t &\sim \mathcal{N}(f^{\text{decoder}}_{\boldsymbol{\mu}}(\boldsymbol{z}_t), \boldsymbol{\Sigma}_{\text{decoder}})
\end{align}
Here $f_{\boldsymbol{\mu}}^{\text{decoder}}$ is the decoder neural network. $\boldsymbol{\Sigma}_{\text{decoder}}$ is a trainable constant covariance matrix for all observations.

We set the prior over $\boldsymbol{z}_t$ to be normally distributed with a trainable constant $\boldsymbol{\mu}_{\boldsymbol{z}}$ and $\boldsymbol{\sigma}_{\boldsymbol{z}}$.
$$\boldsymbol{z}_t \sim \mathcal{N}(\boldsymbol{\mu_z}, \boldsymbol{\sigma_z})$$

Furthermore, for the skill index, we want to find a latent space where the skills are separated in space by directly enforcing the skill switcher to switch based on latent representations. Formally, 
$$\boldsymbol{\delta}_t | \boldsymbol{z}_t \sim \text{Categorical}(f^{\text{switcher}}(\boldsymbol{z}_t))$$
where $f^{\text{switcher}}$ is the switching neural network.

Finally, we assume that the skill policy {\color{red}$f^{\text{policy}}_{\boldsymbol{\delta}_t}(\boldsymbol{\delta}_t, \boldsymbol{z}_t)$} generates control signals by means of a feedback controller acting on the latent states. The control signals are also assumed to follow a Gaussian distribution.
\begin{align*}
    {\color{red}f^{\text{policy}}_{\boldsymbol{\delta}_t}(\boldsymbol{z}_t)} &= p(\boldsymbol{u}_t | \boldsymbol{\delta}_t, \boldsymbol{z}_t) \\
    \boldsymbol{u}_t | \boldsymbol{\delta}_t, \boldsymbol{z}_t &\sim \mathcal{N}(\boldsymbol{k}_t(\boldsymbol{g}_t - \boldsymbol{z}_t), \boldsymbol{\sigma_u}(\boldsymbol{z}_t, \boldsymbol{\delta}_t))
\end{align*}
where $\boldsymbol{g}_t$ and $\boldsymbol{k}_t$ are fixed for a given $\boldsymbol{\delta}_t$. All the goals and gains are internal trainable embeddings. { \color{red}For each time step, given a particular $\boldsymbol{\delta}_t = c$, the goals $\boldsymbol{g}_t$ and gains $\boldsymbol{k}_t$ are extracted from the $c^{\text{th}}$ row of trainable embeddings $\boldsymbol{G}^{C \times S}$ and $\boldsymbol{K}^{C \times S \times A}$ embeddings, where $S$ is the state dimensions and $A$ is the dimension of control signals.}

In summary, for the generation process, given a particular demonstration trajectory $i$, we want to compute and maximize the quantity $p(\boldsymbol{o}_{1:T}, \boldsymbol{u}_{1:T})$, 
\begin{align}
    &p(\boldsymbol{o}_{1:T}, \boldsymbol{u}_{1:T}) = \prod_{t=1}^T p(\boldsymbol{o}_t, \boldsymbol{u}_t)  \label{eq1} \\
           &\quad = \prod_{t=1}^T \int_{\boldsymbol{\delta}_t, \boldsymbol{z}_t} p(\boldsymbol{o}_t|\boldsymbol{z}_t) p(\boldsymbol{u}_t| \boldsymbol{\delta}_t, \boldsymbol{z}_t) p(\boldsymbol{\delta}_t|\boldsymbol{z}_t) p(\boldsymbol{z}_t) d\boldsymbol{z}_t d\boldsymbol{\delta}_t
\end{align}
where
\begin{equation}
    \begin{cases}
        \boldsymbol{z}_t &\sim \mathcal{N}(\boldsymbol{\mu_z}, \boldsymbol{\sigma_z}) \\
        \boldsymbol{o}_t|\boldsymbol{z}_t &\sim \mathcal{N}(f^{\text{decoder}}_{\boldsymbol{\mu}}(\boldsymbol{z}_t), \boldsymbol{\Sigma}_{\text{decoder}}) \\
        \boldsymbol{\delta}_t|\boldsymbol{z}_t &\sim \text{Categorical}(f^{\text{switcher}}(\boldsymbol{z}_t)) \\
        \boldsymbol{u}_t | \boldsymbol{\delta}_t, \boldsymbol{z}_t &\sim \mathcal{N}(\boldsymbol{k}_t(\boldsymbol{g}_t - \boldsymbol{z}_t), \boldsymbol{\sigma_u}(\boldsymbol{z}_t, \boldsymbol{\delta}_t))
    \end{cases}
\end{equation}
It is important to note that conditioned on the goal corresponding to $\boldsymbol{\delta}_t$, the control signal $\boldsymbol{u}_t$ only depends on the current state and transition dynamics are no longer required. In the generation process, we assume that given any random state $\boldsymbol{z}_t$, a state-dependent feedback control law will move the latent state towards an internal goal state. 

\paragraph{Variational Inference}
To approximate the generating process, we use the mean-field approximation family as a proxy for the true posterior distributions, which factors the posterior distribution as $q(\boldsymbol{\delta}_t=c, \boldsymbol{z}_t|\boldsymbol{o}_t, \boldsymbol{u}_t) = q(\boldsymbol{\delta}_t=c|\boldsymbol{z}_t, \boldsymbol{u}_t)q(\boldsymbol{z}_t|\boldsymbol{o}_t)$. We parametrise $q(\boldsymbol{z}_t|\boldsymbol{o}_t)$ using a Gaussian distribution with neural networks that predict the distribution parameters. Conditioned on the latent state $\boldsymbol{z}_t$, we derive analytically the posterior of the goal index,
\begin{align}
    q(\boldsymbol{\delta}_t=c|\boldsymbol{z}_t, \boldsymbol{u}_t) &= \frac{p(\boldsymbol{u}_t|\boldsymbol{\delta}_t=c, \boldsymbol{z}_t)p(\boldsymbol{\delta}_t=c|\boldsymbol{z}_t)}{\sum^C_{c=1} p(\boldsymbol{u}_t|\boldsymbol{\delta}_t=c, \boldsymbol{z}_t)p(\boldsymbol{\delta}_t=c|\boldsymbol{z}_t)} \\
    &= \frac{\mathcal{N}(\boldsymbol{k}_c(\boldsymbol{g}_c-\boldsymbol{z}_t), \boldsymbol{\sigma_u}(\boldsymbol{z}_t, c))\pi(c|\boldsymbol{z}_t)}{\sum^C_{c=1} N(\boldsymbol{k}_c(\boldsymbol{g}_c-\boldsymbol{z}_t), \boldsymbol{\sigma}_u(\boldsymbol{z}_t, c)) \pi(c|\boldsymbol{z}_t)}.
\end{align}

For optimization, we derive the following ELBO \cite{kingma2013auto} loss for this model. See Appendix \ref{Appendix: ELBO} for the full derivation.
{\footnotesize\begin{align}
    & \log p(\boldsymbol{o}_{1:T}, \boldsymbol{u}_{1:T}) \ge \E_{q_{\boldsymbol{z}_t}, q_{\boldsymbol{\delta}_t}}\left[\sum_{t=1}^T \log p(\boldsymbol{o}_t|\boldsymbol{z}_t)\right] + \notag \\
    & \quad \E_{q_{\boldsymbol{z}_t}, q_{\boldsymbol{\delta}_t}} \left[\sum_{t=1}^T\sum_{c=1}^C \mathbbm{1}(\boldsymbol{\delta}_t=c) \log p(\boldsymbol{u}_t|\boldsymbol{\delta}_t=c, \boldsymbol{z}_t)\right] - \notag \\
    & \quad \sum_{t=1}^T KL(q(\boldsymbol{z}_t|\boldsymbol{o}_t), p(\boldsymbol{z}_t)) - \sum_{t=1}^T \E_{q_{\boldsymbol{z}_t}} \left[KL(q(\boldsymbol{\delta}_t|\boldsymbol{z}_t, \boldsymbol{u}_t), p(\boldsymbol{\delta}_t|\boldsymbol{z}_t)) \right] \label{eq2}
\end{align}}
where $\mathbbm{1}(\boldsymbol{\delta}_t=c)$ is the indicator variable which takes the value of 1 when $\boldsymbol{\delta}_t = c$ and 0 otherwise. The $KL$ in the above refers to the Kullback–Leibler (KL) divergence. 

Note that if we sample from $q(\boldsymbol{z}_t|\boldsymbol{o}_t)$, we can calculate the rest of the ELBO loss analytically so that we don't have to sample from the categorical distribution over $\boldsymbol{\delta}_t$. Formally, 

{\footnotesize\begin{align}
    & ELBO = \notag \\
    & \E_{q_{\boldsymbol{z}_t}} \left[ \sum_{t=1}^T \log p(\boldsymbol{o}_t|\boldsymbol{z}_t) +
    \sum_{t=1}^T\sum_{c=1}^C q(\boldsymbol{\delta}_t=c|\boldsymbol{z}_t) \log p(\boldsymbol{u}_t|\boldsymbol{\delta}_t=c, \boldsymbol{z}_t)\right] - \notag \\
    & \sum_{t=1}^T KL(q(\boldsymbol{z}_t|\boldsymbol{o}_t), p(\boldsymbol{z}_t)) - \sum_{t=1}^T \E_{q_{\boldsymbol{z}_t}} \left[KL(q(\boldsymbol{\delta}_t|\boldsymbol{z}_t, \boldsymbol{u}_t), p(\boldsymbol{\delta}_t|\boldsymbol{z}_t)) \right] \label{eq: 14}
\end{align}\par}

\paragraph{Encouraging skill consistency and transition}
Equation \ref{eq: 14} offers an insightful interpretation of the derived ELBO Loss. The first term represents the reconstruction of the observation and control signals based on the posterior samples of $\boldsymbol{z}_t$ and $\boldsymbol{\delta}_t$. The second term is the KL divergence of latent space which ensures the latent space is normally distributed. The third term enforces consistency between the posterior and prior distributions of the switching mechanism.

Furthermore, if we decompose the last term:
\begin{align}
    & \E_{q_{\boldsymbol{z}_t}} \left[KL(q(\boldsymbol{\delta}_t|\boldsymbol{z}_t, \boldsymbol{u}_t), p(\boldsymbol{\delta}_t|\boldsymbol{z}_t)) \right] = \notag \\ 
    & \E_{q_{\boldsymbol{z}_t}} \left[ \mathcal{H}(q(\boldsymbol{\delta}_t|\boldsymbol{z}_t, \boldsymbol{u}_t), p(\boldsymbol{\delta}_t|\boldsymbol{z}_t)) \right]  - \E_{q_{\boldsymbol{z}_t}} \left[\mathcal{H}(q(\boldsymbol{\delta}_t|\boldsymbol{z}_t, \boldsymbol{u}_t))\right]
\end{align}
we see that maximizing the ELBO is equivalent to minimizing the cross-entropy between the prior and posterior distributions of $\boldsymbol{\delta}_t$ for skill consistency. At the same time, it involves maximizing the entropy of the posterior distribution of $\boldsymbol{\delta}_t$, which encourages frequent skill transitions.

In summary, our model shares the same structure as the MDN, and additionally introduces a decoder network that enforces the latent state to contain all the necessary information to reconstruct observations. A key distinction is that the probabilistic structure introduced in our model changes the MDN training objective, which enforces a standard latent state KL divergence term that regularizes the latent space to follow a normal distribution, but also a new switching KL term, which enforces skill consistency. In contrast, the standard MDN training loss only penalises control signal reconstruction errors without imposing any structured probabilistic constraints on the latent space. This structured formulation enables our model to not only predict multimodal outputs like MDNs but also learn a latent state representation that enforces persistent feedback control loops in the network. As shown in the ablation study \ref{sec: ablation study}, this improves robustness to observation noise.

Minimising this objective trains our skill switching and encoder networks, and finds trainable goal and gain embeddings. Once this is complete, we can apply the trained policy in new settings.

\paragraph{Skill Execution}
The detailed skill execution algorithm is described in Algorithm \ref{alg: algorithm 1}. When applying the learned skill at runtime, we first obtain the current latent state by feeding the observations to the encoder (Step 2). Based on this latent state, the skill-switching network generates skill-related goals and gains (Steps 3 and 4). Finally, the latent linear feedback controller outputs the control signals (Step 5). 

As this is a generative model, we can sample different execution paths (Step 2) or choose a maximum likelihood estimate (the mean). Here, we illustrate the sampling approach, which was required to generate unique instances of handwritten characters in the robot handwriting task described in the experimental results below. 

\begin{algorithm}[H]
\caption{Skill Execution algorithm: env environment, switching\_NN switching network}
    \begin{algorithmic}[1]
        \renewcommand{\algorithmicrequire}{\textbf{Input:}}
        \renewcommand{\algorithmicensure}{\textbf{Output:}}
        \REQUIRE $obs$, $model$
        \ENSURE  $u_t$
        \\
        $obs = env.reset()$ \\
        \WHILE{not Done}{
           \STATE $z_t = model.encode(obs).sample() $ \\
           \STATE $c = \arg\max model.switching\_NN(z_t)$  \\
           \STATE $g_t, k_t = lookup([G, K], c)$ \\
           \STATE $u = k_t*(g_t - z_t)$ \\
           \STATE $ obs = env.step(u)$ \\
        }
        \ENDWHILE
    \end{algorithmic}
    \label{alg: algorithm 1}
\end{algorithm}

The derivation and model presented above extend the mixture density network to behave as a switching feedback controller policy in the latent space. Our core hypothesis is that this extension enables the probabilistic combination of feedback controls, where the integration of a probabilistic latent space and switching mechanisms makes the policy more robust and improves skill acquisition.

\section{Evaluation}
\label{sec:result}
The performance of the proposed model is initially evaluated in a realistic and scalable simulation setting, specifically the Franka Kitchen task. We then perform a comprehensive evaluation of the FetchPush task, where we assess its performance and sample efficiency, and conduct an ablation study on various modules of the loss function to investigate the source of the robustness. Lastly, it is deployed on a real UR5 robot to execute writing tasks. 

\begin{table*}[!htb]
    \centering
    \caption{LIBERO Task evaluations: (a) MultiTask success rate: success rate on each of the tasks labelled by their task indices; The number of skills is specified in brackets for each model; (b) Robustness AUC given joint noise: the AUC of success rate on each of the tasks labelled by their task indices; The number of skills is specified in brackets for each model;}
    \subfloat[Success Rate on each task for LIBERO]{
        \begin{tabularx}{\linewidth}{lXXXXXXXXXXX}
        \toprule
             \textbf{Task Number} & \textbf{1} &  \textbf{2} &  \textbf{3} &  \textbf{4} & \textbf{5} & \textbf{6} & \textbf{7} & \textbf{8} & \textbf{9} & \textbf{10} & \textbf{Avg} \\
            \midrule
             \textbf{MDN (5)} & 70\% &  80\% & 65\% & 40\% & 70\% & 60\% & 60\% & 90\% & 80\% & 75\% & \textbf{69\%} \\
        
             \textbf{Our Model (5)} & 80\% & 100\% & 90\% & 40\% & 100\% & 60\% & 90\% & 100\% & 85\% & 80\% & \textbf{83\%} \\
             \bottomrule
        \end{tabularx}
        \label{tab: Libero performance}
    }%
    \hfil
    \subfloat[AUC on each task given sensory noise in LIBERO]{
        \begin{tabularx}{\linewidth}{llXXXXXXXXXXX}
        \toprule
        \textbf{Noise Type} & \textbf{Task Number} & \textbf{1} &  \textbf{2} &  \textbf{3} &  \textbf{4} & \textbf{5} & \textbf{6} & \textbf{7} & \textbf{8} & \textbf{9} & \textbf{10} & \textbf{Avg} \\
        \midrule
        \multirow{2}{*}{\textbf{Joint Noise}} & \textbf{MDN (5)} & 84\% & 76\% & 54\% & 37\% & 72\% & 66\% & 59\% & 99\% & 65\% & 57\% & \textbf{67\%} \\
        & \textbf{Our Model (5)} & 74\% & 98\% & 66\% & 43\% & 90\% & 56\% & 79\% & 94\% & 82\% & 73\% & \textbf{75\%} \\
        \bottomrule
        \end{tabularx}
        \label{tab: libero robustness}
        }%
\end{table*}

\subsection{Metrics}
\subsubsection{Success Rate} We use the success rate as the performance metric for the models, which measures the number of successful trials in the test set defined by each of the simulated environments. For the physical experiment, we perform a qualitative analysis based on both the simulation and real-world deployment.

\subsubsection{Robustness AUC} To measure the robustness, we add different levels of noise to the observations and monitor the success rates. We use the AUC (Area Under the Curve) of those success rates given different levels of noise as the robustness metric.

Specifically, we first collect the standard deviation of the input from the training dataset by averaging across trajectories and time steps. We introduce varying levels of noise by sampling from a Normal distribution with different standard deviation scales and adding these to the observations. The performance is evaluated by calculating the average success rate in the test dataset.

\subsubsection{Skill Transitions} We report the number of skills utilized and the skill duration, defined as the average fraction of time steps spent
consecutively executing a given skill during a task. The larger the fraction is, the longer a skill executes.

\subsection{Franka Kitchen Task}
\subsubsection{Dataset} For the franka kitchen task, we use the LIBERO benchmark \cite{liu_libero_2023}.  LIBERO offers a robust and scalable evaluation framework with four task suites: LIBERO-SPATIAL, LIBERO-OBJECT, LIBERO-GOAL, and LIBERO-100. We only use the LIBERO-GOAL datasets for our evaluation. These contain demonstrations of robot arms moving to different places and interacting with various objects in a fixed scene. The observations for this task are multimodal and include three types of input: visual observations (workspace view and view in hand, both are 128x128x3 RGB images), proprioceptive observations (end effector, joint and gripper states) and task descriptions in natural language form. The task goals are specified using a propositional formula that aligns with the language instructions \cite{liu_libero_2023}. The action space is a 7-dimensional joint angle control signal. We use the LIBERO-goal dataset shown in Figure \ref{fig: libero Dataset} to learn goals and gains that can be reused across different tasks. There are 10 tasks in total and for each task, there are 50 demonstrations of various lengths. Modelling such tasks requires the model to extract task-relevant information from large language model embeddings and to align visual observations of the environment with the agent's proprioceptive states in order to navigate robustly
\subsubsection{Baseline} For the Mixture Density Network (MDN) baseline, we choose the existing LIBERO state-of-the-art ResNet-T \cite{liu_libero_2023} model. Our model is identical, with the exception of the action head. We choose the "multitask" setting to train baselines, randomly sampling from multitask data rather than sequentially. More details can be found in \cite{liu_libero_2023}.
\subsubsection{Analysis}
\begin{figure}
    \centering
    \includegraphics[width=\linewidth, trim={1cm 0cm 2cm 1cm}, clip]{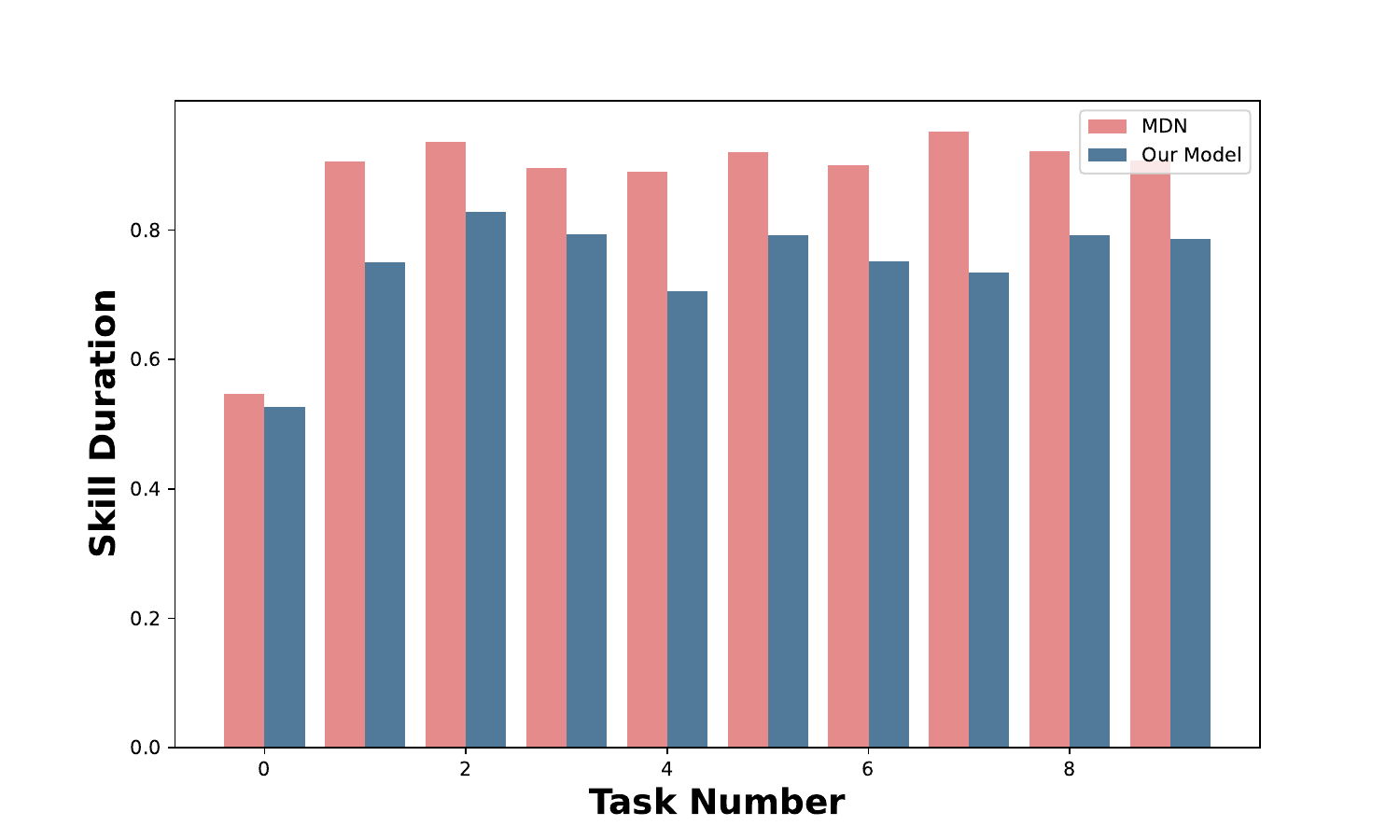}
    \caption{Skill Duration for Each Task: The averaged fraction of time-steps spent consecutively executing a skill during a task. Our model executes skills for shorter durations and relies less on any single skill than the MDN.}
    \label{fig: skill duration}
\end{figure}

Firstly, based on the perfect demonstrations, our model improves upon the MDN by 12\% on average, as shown in Table \ref{tab: Libero performance}. Our model achieves an equivalent or better success rate on every task. Secondly, to evaluate the robustness of our model, we apply different noise scales of 
$[0, 0.1, 0.2, 0.3, 0.5, 1, 2, 3]$ and calculate the Robustness AUC in Table \ref{tab: libero robustness}. Our model is also better than the MDN by 8\% on average. Finally, as shown in Figure \ref{fig: skill duration} and Table \ref{tab: skill summary}, our model prefers more skills and more skill transitions in line with expectations based on the additional KL term in our loss. This results in a model that executes skills for shorter durations and relies less on any single skill than the MDN. 

Our experimental results demonstrate that our approach significantly outperforms the standard MDN baseline with or without sensory noise. This performance gain highlights the scalability and robustness of the model in handling complex, multimodal tasks.

\begin{table}[!htb]
    \centering
    \caption{Descriptive Statistics for skill transtions}
    \begin{tabularx}{\columnwidth}{lXX}
    \toprule
     Models & MDN  &  Our Model \\
     \midrule
     The number of used skills per task & $4.8 \pm 0.6$ & $5 \pm 0$ \\
     Avg skill duration per task & $0.88\pm0.11$ & $0.75\pm0.08$ \\
     Avg transition fraction per task & $0.12 \pm 0.11$ & $0.25\pm0.08$\\
     \bottomrule
    \end{tabularx}
    \label{tab: skill summary}
\end{table}

Since our model produces the skill index for each step, we gain some level of interpretability for each skill. We report skill interpretation results based on topic modelling in Appendix \ref{Appdenix: skill interpretation}.

\subsection{FetchPush Task}
\subsubsection{Dataset} The FetchPush  \cite{towers_gymnasium_2023} task requires a robot manipulator to move a block to a target position on top of a table by pushing with its gripper, as shown in Figure \ref{fig: fetchpush}. The task is marked as a success when the block is moved to the target within 50 time steps. The robot is a 7-DoF Fetch Mobile Manipulator with a two-fingered parallel gripper. The robot is controlled by commanding small displacements of the gripper in 3D Cartesian coordinates, with robot inverse kinematics computed internally by the MuJoCo framework \cite{todorov2012mujoco}. The gripper is locked in a closed configuration to perform the push task. The task continues for a fixed number of time steps, which means that the robot has to maintain the block in the target position once reached. The observation is a goal-aware observation space. It consists of a 25-dimensional vector including the state and velocity of the robot and the block and 3-dimensional Cartesian positions of the goal. The action space is the Cartesian displacement dx, dy, and dz of the end effector. We use pre-trained RL agents from rl-zoo3 \cite{rl-zoo3} and stablebaselines3 \cite{stable-baselines3} to collect 9800 expert demonstration trajectories. 
\subsubsection{Baselines}  We implement two baseline policies for the FetchPush task, Behaviour Cloning (BC) and MDN. For BC, we use the same structure as the behaviour cloning baseline from \cite{lee_generalizable_2021}, an MLP with 2 hidden layers of 256 hidden units as the feature extractor and a linear layer to transform 256 hidden units to control signals. For the MDN, we implement the mixture density network using the same skill-switching neural network as our model, and the encoder as an action predictor that outputs actions depending on the predicted skill index.

\begin{table*}[!htb]
    \centering
    \caption{Average success rate on test seed based on 5 trained models given each setting in FetchPush}
    \subfloat[FetchPush: Average Success Rate given different numbers of skills]{
    \begin{tabularx}{\linewidth}{lXXXX}
        \toprule
        \textbf{Number of Skills} & \textbf{5} & \textbf{10} & \textbf{20} & \textbf{100}\\
        \midrule
        \textbf{MDN} & \textbf{79.08\%$\pm$2.70\%} & \textbf{84.09\%$\pm$1.84\%} & 77.25\%$\pm$3.54\% & 69.00\%$\pm$3.30\% \\
        \textbf{Our model} & 74.40\%$\pm$3.19\% & 82.00\%$\pm$2.61\% & \textbf{84.40\%$\pm$2.86\%} & \textbf{82.80\%$\pm$1.96\%} \\
        \bottomrule
    \end{tabularx}
    \label{fig: average success rate skill number}
    }%
    \quad
    \subfloat[FetchPush: Average Success Rate given different training data size with the optimal skill number]{
        \begin{tabularx}{\linewidth}{lXXXX}
            \toprule
            \textbf{Sample Size} & \textbf{25\%} & \textbf{50\%} & \textbf{75\%} & \textbf{100\%} \\
            \midrule
            \textbf{BC} &  59.09\%$\pm$2.55\% & 74.45\%$\pm$4.61\% & 79.25\%$\pm$3.10\% & 83.25\%$\pm$1.89\% \\
            \textbf{MDN (10)} &  62.40\%$\pm$4.45\% & \textbf{74.80\%$\pm$2.42\%} &  76.80\%$\pm$2.73\% &  84.09\%$\pm$1.84\% \\
            \textbf{Our model (20)} &  \textbf{74.40\%$\pm$4.79\%} &  74.40\%$\pm$2.93\% & \textbf{82.40\%$\pm$2.14\%}	& \textbf{84.40\%$\pm$2.86\%} \\
            \bottomrule
        \end{tabularx}
    \label{fig: average success rate sample size}
    }%
    \label{fig: average success rate}
\end{table*}

\subsubsection{Analysis}
\paragraph{Task Performance} Table \ref{fig: average success rate skill number} provides the average success rates of different models with different skill numbers and training data sizes. As the number of skills increases, our model's performance improves significantly, outperforming the MDN when there are 20 and 100 skills, but underperforming when there are 5 or 10 skills. This suggests that MDN can achieve comparable performance to ours with enough training samples and the optimal number of skills. 

\paragraph{Robustness} As shown in Figure \ref{fig: Obs noise}, We plot the success rate curve of each model at its best number of skills, given different levels of noise. Firstly, our model performs significantly better compared with other models across all different noise levels. Secondly, we note that our model has no performance loss for up to 5\% added noise. The AUC of our model decreases from 84\% to 80\% for 10\% added noise. For 20\% of added noise, the success rate of our model remains around 58\%, whereas the other models are around or below 45\%. This confirms our hypothesis that the proposed switching feedback controller structure results in more robust control.

\begin{figure}[!htb]
    \definecolor{color1}{rgb}{0.80851064, 0.09219858, 0.09929078}
    \definecolor{color2}{rgb}{0.15068493, 0.34520548, 0.50410959}
    \definecolor{color3}{rgb}{0.23619632, 0.53680982, 0.22699387}
    \definecolor{color4}{rgb}{0.38676845, 0.19847328, 0.41475827}
    \centering
        \begin{tikzpicture}
            \begin{axis}[
                xlabel={Noise level}, 
                ylabel={Success Rate},
                axis lines=middle,
                ymajorgrids=true,
                grid style={dashed},
                width=0.8\linewidth,
                height=0.5\linewidth,
                xmin=-0.0,
                xmax=0.32,
                ymin=0,
                ymax=1,
                clip=false,
                xticklabel style={/pgf/number format/.cd,fixed,precision=2},
                legend style={
                at={(0.25, 0.4)},
                anchor=north, 
                nodes={scale=0.7, transform shape}
                }
            ]
            \addplot+[
              no markers,
              color=color1,
              line width=2pt,
              opacity=.7,
              error bars/.cd, 
                y fixed,
                y dir=both, 
                y explicit,
            error bar style={line width=2pt,solid}
            ] table[x=noise, y=suc,y error=error]{
            noise   suc error
            0.0   0.832000  0.041473  5.0
            0.05  0.716000  0.062290  5.0
            0.1   0.676000  0.062290  5.0
            0.15  0.580000  0.034641  5.0
            0.2   0.444000  0.060663  5.0
            0.25  0.356000  0.049800  5.0
            0.3   0.264000  0.065422  5.0
            };
            \addlegendentry{BC};
            \addplot+[
              color=color2,
              no markers,
              line width=2pt,
              opacity=.7,
              error bars/.cd, 
                y fixed,
                y dir=both, 
                y explicit,
              error bar style={line width=2pt,solid}
            ] table[x=noise, y=suc,y error=error]{
            noise   suc error
            0.0   0.840000  0.040000  5.0
            0.05  0.808000  0.046043  5.0
            0.1   0.708000  0.038987  5.0
            0.15  0.584000  0.045607  5.0
            0.2   0.472000  0.080747  5.0
            0.25  0.324000  0.038471  5.0
            0.3   0.312000  0.070143  5.0
            };
            \addlegendentry{MDN (10)};
            \addplot+[
              no markers,
              color=color4,
              line width=2pt,
              opacity=.7,
              error bars/.cd, 
                y fixed,
                y dir=both, 
                y explicit,
             error bar style={line width=2pt,solid}
            ] table[x=noise, y=suc,y error=error]{
            noise   suc error
            0.0   0.844000  0.063875  5.0
            0.05  0.860000  0.044721  5.0
            0.1   0.808000  0.038987  5.0
            0.15  0.680000  0.050990  5.0
            0.2   0.580000  0.060000  5.0
            0.25  0.404000  0.071274  5.0
            0.3   0.340000  0.092736  5.0
            };
            \addlegendentry{Our Model (20)};
            \end{axis}
            \end{tikzpicture}
    \caption{FetchPush Robustness Curve: The red and blue lines are the Success Rate curve of baselines of optimal skill number, BC and MDN respectively. The purple line is our model of optimal skill number. We report the average success given different noise levels for each model.}
    \label{fig: Obs noise}
\end{figure}

\begin{table*}[!htb]
    \centering
    \caption{Robustness calculated on the AUC of the Success Rate given different factors.}
    \subfloat[FetchPush: Robustness given different skill number]{
    \begin{tabularx}{\linewidth}{lXXXX}
        \toprule
        \textbf{Number of Skills} & \textbf{5} & \textbf{10} & \textbf{20} & \textbf{100}\\
        \midrule
        \textbf{BC} & 55.26\%$\pm$3.61\% & 55.26\%$\pm$3.61\% & 55.26\%$\pm$3.61\% & 55.26\%$\pm$3.61\%\\
        \textbf{MDN} & 57.47\%$\pm$3.05\% & 57.83\%$\pm$2.21\% & 56.46\%$\pm$5.86\% & 53.49\%$\pm$4.20\% \\
        \textbf{Our model} & \textbf{58.57\%$\pm$4.30\%} & \textbf{62.17\%$\pm$5.64\%} & \textbf{64.51\%$\pm$3.74\%} & \textbf{63.60\%$\pm$4.72\%} \\
        \bottomrule
    \end{tabularx}
    \label{fig: robustness tables given skill}
    }%
    \hfil
    \subfloat[FetchPush: Robustness given different training data size]{
        \begin{tabularx}{\linewidth}{lXXXX}
            \toprule
            \textbf{Sample Size} & \textbf{25\%} & \textbf{50\%} & \textbf{75\%} & \textbf{100\%} \\
            \midrule
            \textbf{BC} & 40.46\%$\pm$2.56\% & 49.77\%$\pm$7.85\% & 54.34\%$\pm$5.19\% & 55.26\%$\pm$3.61\% \\
            \textbf{MDN (10)} & 43.94\%$\pm$5.01\% & 50.69\%$\pm$2.63\% & 55.09\%$\pm$4.47\% & 55.09\%$\pm$5.77\% \\
            \textbf{Our model (20)} & \textbf{53.66\%$\pm$3.55\%} & \textbf{54.51\%$\pm$4.32\%} & \textbf{60.34\%$\pm$3.16\%} & \textbf{64.51\%$\pm$3.74\%} \\
            \bottomrule
        \end{tabularx}
        \label{fig: robustness tables given sample size}
    }
    \label{tab: robustness tables}
\end{table*}

\paragraph{Sample efficiency}
As shown in Table \ref{fig: average success rate sample size}, our model achieves better performance given fewer samples. As the sample size increases, our model consistently performs well, matching or exceeding the success rates of the MDN and BC at 50\%, 75\%, and 100\% of the training data. In addition, our model with 25\% samples achieves similar performance as one with 50\%. 

In addition, we calculate the robustness AUC of each model for the different numbers of skills and different training data sample sizes shown in Table \ref{tab: robustness tables}. Our model significantly outperforms others in every skill and sample setting. Secondly, we observe that with the increase in the sample size (Table \ref{fig: robustness tables given sample size},), the performance increases consistently for all models, which suggests that an increase in training data improves the model performance. Thirdly, our model significantly outperforms others in every sample size setting, which demonstrates its effectiveness across different training data sizes.

\paragraph{Ablation Study}
\label{sec: ablation study}
For the ablation study, we focus on understanding the source of the model’s robustness. As shown in Table~\ref{fig: skill number ablation}, we evaluate the impact of different components of the loss function on performance under noise. The \textbf{MDN} baseline corresponds to a standard mixture-of-experts formulation without any structural constraints. \textbf{MDN + FB} incorporates the feedback controller structure into the MDN, resulting in an approximate 3\% improvement in robustness. \textbf{MDN + FB + SW} further introduces skill-switching consistency, which enables more effective transitions between skills and shows additional robustness benefits.

While frequent skill transitions are not directly tied to robustness, we hypothesize that they encourage more complex and blended behaviours. Such behaviour may help reduce reliance on individual skills and improve the model’s ability to generalize across varied conditions, thereby contributing indirectly to overall robustness. These results highlight the contribution of both feedback-based control structures and skill-switching dynamics in improving model performance.

The underlying reasons why the feedback controller design contributes to robustness are further discussed in Section~\ref{sec: discuss of source of robustness}, where we provide additional analysis and empirical evidence to support this hypothesis.

\begin{table*}[!hbt]
    \centering
    \caption{Ablation study with different modules: FB, feedback controller desigan; SW, switching KL divergence; 
    }
    \subfloat[Ablation Study given different skills]{
        \begin{tabularx}{\linewidth}{lXXXX}
            \toprule
            \textbf{Model} & \textbf{5} & \textbf{10} & \textbf{20} & \textbf{100} \\
            \midrule
            \textbf{MDN} & 57.47\%$\pm$3.05\% & 57.83\%$\pm$2.21\% & 56.46\%$\pm$5.86\% & 53.49\%$\pm$4.20\% \\
            \textbf{MDN, FB} & 55.14\%$\pm$3.46\% & 60.44\%$\pm$3.51\% & 61.03\%$\pm$3.03\% & 62.53\%$\pm$5.52\% \\
            \textbf{MDN, FB, SW} & 58.57\%$\pm$4.30\% & 62.17\%$\pm$5.64\% & \textbf{64.51\%$\pm$3.74\%} & 63.60\%$\pm$4.72\% \\
            \bottomrule
        \end{tabularx}
        \label{fig: skill number ablation}
    }%
\end{table*}

\begin{figure*}[!htb]
    \centering
    \subfloat[Robot Writing task]{
    \includegraphics[width=0.97\columnwidth, trim={8cm 3cm 8cm 3cm},clip]{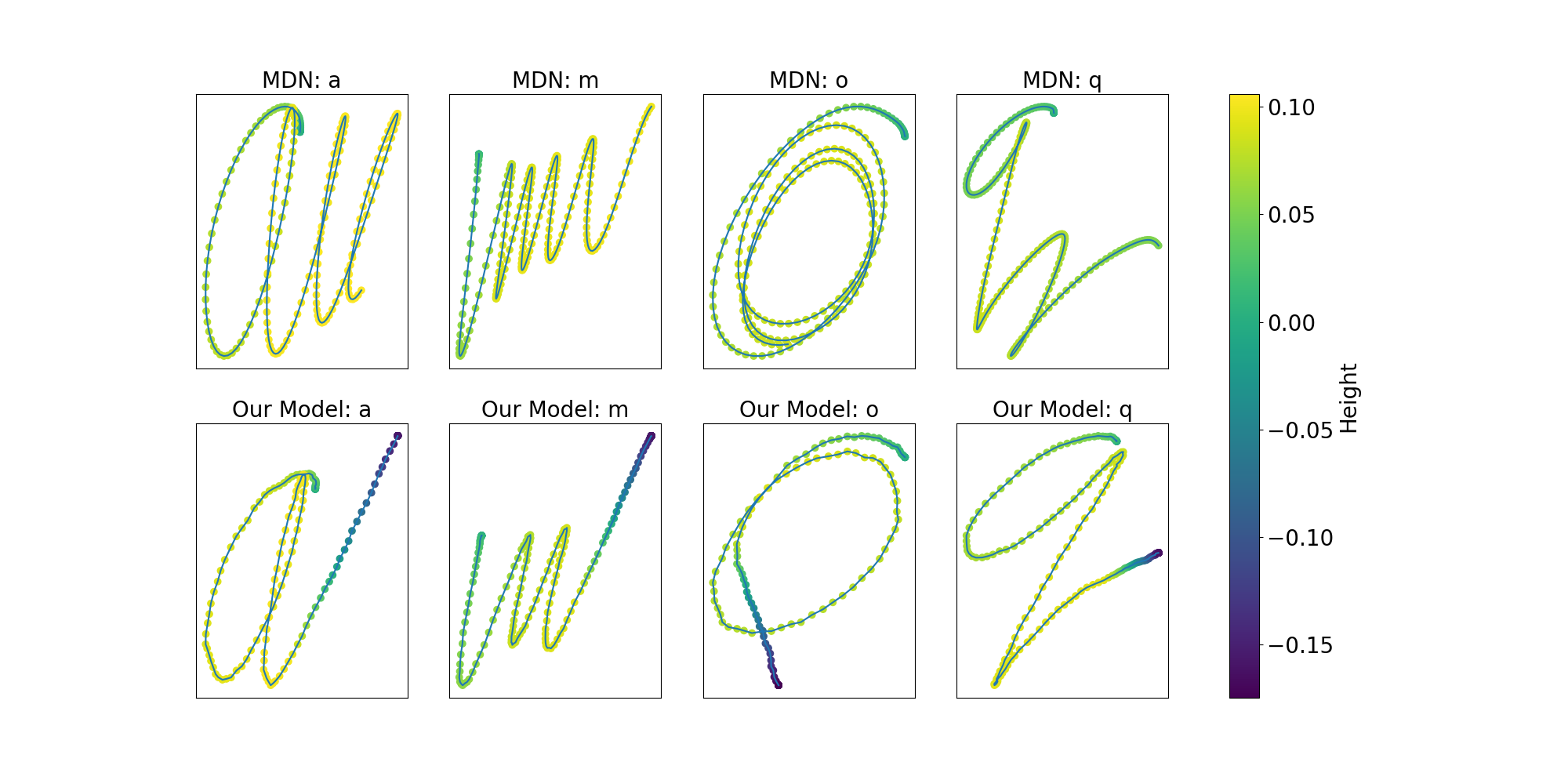}
    }
    \hfil
    \subfloat[Physical deployment]{
        \includegraphics[width=0.85\columnwidth, trim={0cm 0cm 4cm 0cm}, clip]{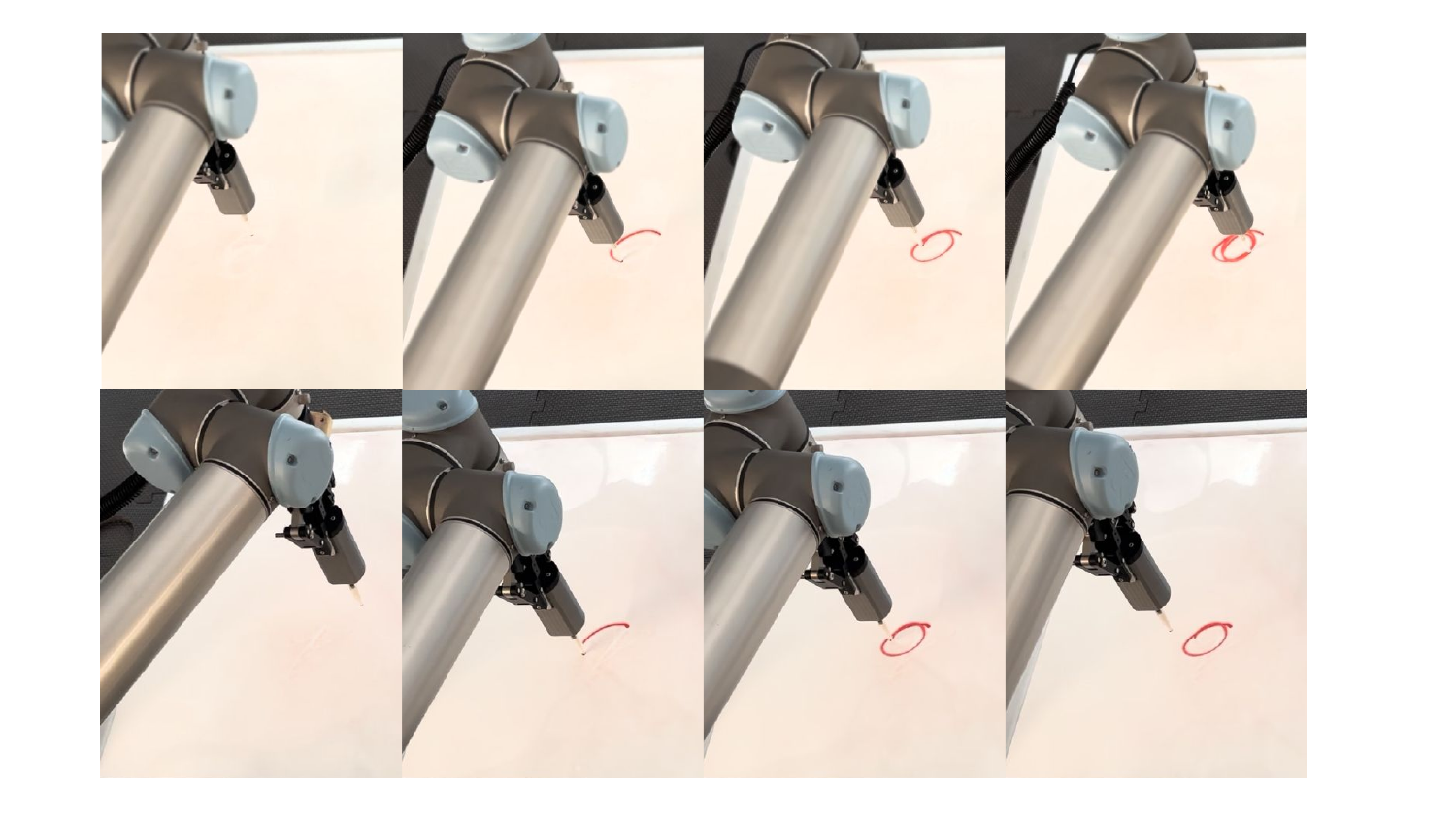}
    }
    \caption{Robot Writing Task: (a) trajectories predicted by different models in simulation; The first row shows trajectories generated by the MDN; The second row shows the trajectories generated by our model. Positive height corresponds to writing pressure when in contact with the whiteboard, while negative height indicates a pen lift. (b) Physical deployment of different models: The first row shows control with the MDN model; The second row shows latent feedback control with our model. Our model stops appropriately after the letter o is finished, while the MDN keeps drawing the letter.}
    \label{fig: writing task}
\end{figure*}

\subsection{Robot Writing Task}
\subsubsection{Dataset} The Character Trajectories dataset \cite{character_trajectories_175} consists of multiple labelled samples of human pen tip movements on a tablet, recorded while writing individual characters. All samples come from the same writer writing letters in different ways and are intended for primitive extraction. Only characters with a single pen-down segment are included, covering a total of 20 letters and 2858 trajectories. Each trajectory state is represented in 3D Cartesian space, and the actions correspond to the displacements in the same space. We set up a robot character writing task using this dataset, which requires the robot to learn to write each of the letters based on the demonstrated human writing trajectories. We train models on all demonstrations, using a character embedding token to select a desired character to be written.
\subsubsection{Baselines} For the MDN baseline, we use a 3-layer MLP combined with an LSTM as the state encoder and action head as a single-layer MDN for the switching feedback controllers. Our model only differs in the action head. We deploy the model on a UR5 robot. During testing, we compute the immediate waypoint with the current pose from the sensor and the predicted displacement from the model. A position controller is then used to guide the UR5 based on the waypoints. To ensure smoother execution, the immediate waypoint is generated when the robot is within a certain distance of the previous waypoint, taking advantage of the model’s robustness to observation noise.
\subsubsection{Analysis} 
\label{sec: robochar analysis}
Our model is able to generate variations of all the letters smoothly by sampling the latent states (see Algorithm \ref{alg: algorithm 1} for details). A video demonstration can be found in the supplementary material. As illustrated in Figure \ref{fig: writing task}, the model appropriately stops at the end of the task, thanks to our explicit feedback controller design, which leads to stable convergence to a latent set-point or goal. In contrast, the MDN produces unstable displacements towards the end of the drawing, which makes it difficult to deploy, as it is unclear when the character drawing task is complete.

We visualize the trajectory with skill indices for our model and MDN in the Appendix \ref{Appendix: robot writing}. Our model utilizes a wider range of skills, resulting in improved letter trajectories.

\subsection{Robustness and Stability}
\label{sec: discuss of source of robustness}

The latent feedback control structure naturally lends itself to stability analysis, assuming linear dynamics models can be fit in the latent space corresponding to each controller. Assuming
linear dynamics 
\begin{equation}
    \mathbf{z}_t = \mathbf{A}^i\mathbf{z}_{t-1} + \mathbf{B}^i\mathbf{u}_t
\end{equation} in the latent space corresponding to the $i$-th feedback controller, the closed loop system is stable if the real parts of the eigenvalues of the closed loop system ($\mathbf{A}^i+\mathbf{B}^i\mathbf{K}^i$) are negative \cite{bishop2011modern}. Prior work on Koopman operators has explored approaches that learn stable locally linear dynamics models \cite{brunton_koopman_2016, shi_deep_2022} that enforce this directly. Unfortunately, this is challenging to obtain in an imitation learning setting where we need to learn from only a few closed-loop human demonstrations, particularly when we are operating on high-dimensional inputs such as images or text (our libero experiments). Given the need to exhaustively explore state transitions to learn a good dynamics model, dynamics modelling may not be realistic for this class of problems. 

\begin{figure*}[!htb]
    \centering
    \includegraphics[width=\linewidth, trim={11cm 6cm 19cm 11cm}, clip]{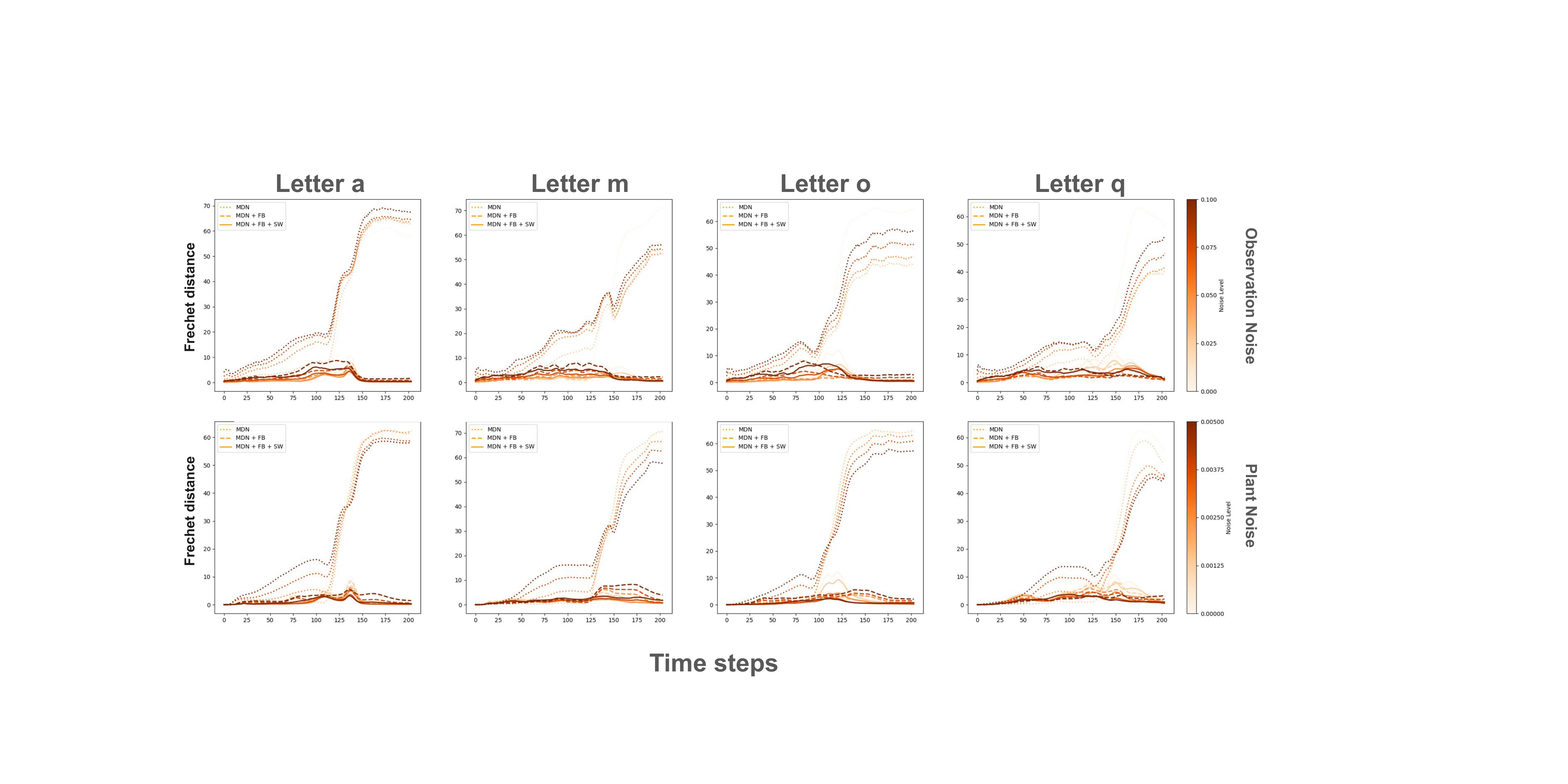}
    \caption{Fréchet distance between latent spaces of generated trajectories and training dataset trajectories under varying levels of observation noise (top row) and plant noise (bottom row) for four different tasks (a, m, o, q). Lower distances indicate better trajectory distribution matching. Models incorporating feedback controller structure (MDN + FB and MDN + FB + SW) consistently achieve lower distances compared to the standard MDN baseline, indicating improved robustness under noisy conditions. MDN baselines diverge from the training set as these models go unstable during execution, while models with feedback structures keep latent states bounded near the distribution of latent states seen at training time.}
    \label{fig:fd_noise_robustness}
\end{figure*}

Instead, for imitation learning settings, we propose to conduct an empirical stability-like analysis and robustness test. We illustrate this for the robot writing task. Here, we evaluate the model's behaviour under varying levels of noise, as shown in Figure~\ref{fig:fd_noise_robustness}, and use the Fréchet Distance as a measure of the deviation between latent states of generated and demonstrated trajectories. Intuitively, this measures whether generated trajectories under the model lie within the distribution of those seen at training time. An unstable controller would likely result in latent states that diverge from the training set, while a stable controller would keep generating states bounded close to those seen at training time. The plots compare the Fréchet Distance between model-deployed trajectories and ground truth trajectories across different time steps under increasing levels of observation and plant noise. During deployment, our model consistently produces actions that remain closer to the ideal trajectory and exhibit empirically bounded deviation at the end of the trajectories, whereas the standard MDN baseline suffers from cumulative error and increasing divergence from the target distribution as it goes unstable during execution.

Across all tasks and noise conditions, models with feedback controller structures (MDN + FB and MDN + FB + SW) consistently achieve lower Fréchet distances when compared to the MDN baseline. This supports the hypothesis that the feedback controller design contributes directly to improved robustness by promoting more stable and consistent behaviour in noisy environments.

While the underlying mechanism has not been rigorously proven, our empirical results strongly suggest that the feedback controller provides an implicit regularization effect by anchoring the policy closer to the training data distribution. In particular, the explicit feedback control structure, comprising learned goals and gains, continuously guides the system toward target goal states, making it less prone to diverge under perturbations. This mechanism effectively constrains trajectories within the distribution of behaviours observed during training. The goal-directed behaviour also contributes to improved task completion properties: The model tends to stabilize and stop after completing the writing rather than drifting unnecessarily, as often observed with the standard MDN baseline (\autoref{fig: writing task}). Such stabilization not only enhances robustness but also results in more interpretable and desirable behaviour in practical applications.

\section{Discussion}
\label{sec:Discussion}
Our approach demonstrates significant improvement in robustness against observation noise, highlighting the potential of integrating classical control theories into modern neural network frameworks. Considering that different initialization methods \cite{he2015delving, glorot2010understanding} yield weights with varying magnitudes of eigenvalues in the context of linear neural networks, this opens up the possibility of exploring classical controllers with different gains in the latent space to improve robustness.

It is important to note that the latent feedback controller structure proposed here may not segment tasks into linear sequences of skills that directly align with human expectations. For example, in the Libero test environment, for some tasks, we see rapid switching between skills to generate desired behaviours, while in others, we see clearer sequencing of skills executed for longer periods of time. For the former, a latent goal may not necessarily need to be reached to generate a desired behaviour, and one or more goals could be sequenced to achieve some desired control outcome. Moreover, the same skill may generate different behaviours when executed in different parts of the latent space. For example, in the character drawing task, while we see clear skills for pen down, lifting and stopping, character drawing itself is primarily executed by a single skill, with different trajectories generated by the same feedback control law, clearly by following different paths through the latent space towards the skill's latent goal. In this case, it appears that the model has found a latent state manifold that is rich enough to generate a broad range of behaviours using only a single feedback control law.

Our approach is not without limitations. One limitation is that the number of skills in the model is a meta-parameter that needs to be predetermined before training. A lower number of skills can lead to degenerate task performance due to insufficient capacity, while a higher number may result in overly fragmented or redundant segmentations. Although our current formulation requires the number of skills to be specified in advance, this is not a fundamental scalability bottleneck. In practice, different skill numbers can be evaluated in parallel, and the best-performing configuration can be selected based on validation performance and available computational resources.

Moreover, our model consistently improves the robustness and sample efficiency of the MDN baseline across different numbers of skills and training sample sizes. While our ablation study explores a range of skill configurations with sufficient demonstrations, real-world scenarios often lack the time, data, or computational resources to exhaustively search for the optimal number of skills. In this context, our experiments show that simply incorporating our model into an MDN, using a suboptimal skill number and few demonstrations, still leads to improved performance and robustness. This highlights the practicality of our approach as a plug-and-play enhancement for MDN-based systems, offering reliable gains even when skill configurations are not carefully tuned.

To further reduce the need for manual tuning, future work could explore incorporating priors such as Dirichlet processes. These nonparametric regularizers would allow the model to infer the number of necessary skills automatically, enabling flexible model capacity, better generalization, and support for lifelong skill acquisition.

Another limitation is that transitions between skills are determined by random sampling. High-level requirements that guide these transitions are missing, which could help achieve more coherent skill sequences. 

In addition, since the learned latent space is entirely dependent on the model, it is possible for consecutive real states to be mapped to distant locations within the latent space. Introducing a latent dynamics process (eg. \cite{jaques_newtonianvae_2021, brunton_koopman_2016, shi_deep_2022}) to enforce smoother transitions in the latent space could be beneficial.

Finally, while our formulation is agnostic to the specific type of control signal, it is most naturally suited to velocity and positional control. Extending the approach to torque control is theoretically feasible \cite{gao2020learning, herzallah2004mixture}, but in practice, it presents additional challenges such as increased sensitivity to system dynamics and the need for higher-frequency inference. Moreover, although our model jointly learns both gain and goal parameters, there may be representational redundancy in this formulation—for example, multiple skills may share similar goals or different gain–goal pairs may yield similar behaviours. In the context of a shared latent space and a skill-switching mechanism, the former case can be seen as a natural consequence of our parameterization. This flexibility can be advantageous as it allows the model to represent a diverse range of behaviours and adapt dynamically across different phases of a task.

The latter case, however, may result in multiple controllers explaining the same behaviour, a phenomenon related to mode collapse in mixture models. To mitigate this, we adopt a design principle similar to that studied in \cite{dilokthanakul2016deep} and further discussed in the follow-up\footnote{The M2 model blocks the independent path of $z$ to reconstruct observations $x$. See: \url{https://ruishu.io/2016/12/25/gmvae/}}. Specifically, we prevent the model from bypassing the categorical latent variable via alternative probabilistic paths, encouraging it to make effective use of discrete skill assignments and maintain meaningful segmentation. Empirically, we observe that gain–goal combinations are activated differently depending on task context and skill transitions, suggesting that the model effectively leverages its expressive capacity. While this form of redundancy has not caused practical issues in our implementation, future work could explore structured regularization of gain matrices or incorporate control-theoretic constraints—such as those inspired by Koopman operator theory \cite{brunton_koopman_2016, shi_deep_2022} —to learn more interpretable controller structures.

\section{Conclusion}
In this work, we derived a probabilistic graphical model to describe the skill acquisition process as segmentation in the latent space, with each skill policy functioning as a feedback control law. Our approach demonstrated significant improvements in robustness against observation noise and sample efficiency, showcasing the potential of integrating classical control theories into modern neural network frameworks. A notable finding from our research is the behaviour of the controller in the latent state space. Unlike traditional models where states are fixed and predefined, our approach uses learnable latent states. This adaptability allows the controller to dynamically adjust its behaviour based on the learned states, potentially leading to more efficient and effective control policies. Future research could continue to explore the potential of latent representations and refine the integration of classical control theories with modern neural network frameworks to further enhance the robustness and efficiency of models.

\section*{Acknowledgments}
This research is supported by the Monash Graduate Scholarship. We are grateful to the members of Monash Robotics for general suggestions, especially Rhys Newbury and Tin Tran, for valuable discussions and recommendations around robot deployment.

\appendices
\section{Proof of the ELBO Loss}
\label{Appendix: ELBO}
First, we assume each observation and action pair is independent of each other.
$$p(\boldsymbol{o}_{1:N, 1:T}, \boldsymbol{u}_{1:N, 1:T}) = \prod_{i=1}^N \prod_{t=1}^T p(\boldsymbol{o}_{it}, \boldsymbol{u}_{it})$$
For simplicity, we ignore sample index $i$, for each pair $(\boldsymbol{o}_t, \boldsymbol{u}_t)$, 
{\footnotesize\begin{align}
    & \log p(\boldsymbol{o}_t, \boldsymbol{u}_t) \ge \notag \\
    & \E_{q_{\boldsymbol{z}_t}, q_{\boldsymbol{\delta}_t}}\left[\log\frac{p(\boldsymbol{o}_t|\boldsymbol{z}_t)p(\boldsymbol{z}_t) \prod_{c=1}^C \left(p(\boldsymbol{u}_t|\boldsymbol{\delta}_t, \boldsymbol{z}_t)p(\boldsymbol{\delta}_t|\boldsymbol{z}_t) \right)^{\mathbbm{1}(\boldsymbol{\delta}_t=c)}}{q(\boldsymbol{z}_t|\boldsymbol{o}_t)\left(q(\boldsymbol{\delta}_t|\boldsymbol{o}_t, \boldsymbol{u}_t)\right)^{\mathbbm{1}(\boldsymbol{\delta}_t = c)}} \right] \\
    &= \E_{q_{\boldsymbol{z}_t}}\left[\log p(\boldsymbol{o}_t|\boldsymbol{z}_t) \right] + 
    \E_{q_{\boldsymbol{z}_t}, q_{\boldsymbol{\delta}_t}}\left[\sum_c \mathbbm{1}(\boldsymbol{\delta}_t = c) \log p(\boldsymbol{u}_t|\boldsymbol{\delta}_t = c, \boldsymbol{z}_t) \right] + \notag \\
    & \E_{q_{\boldsymbol{z}_t}}\left[\log \frac{p(\boldsymbol{z}_t)}{q(\boldsymbol{z}_t|\boldsymbol{o}_t)}\right] + 
    \E_{q_{\boldsymbol{z}_t}, q_{\boldsymbol{\delta}_t}}\left[\log\frac{p(\boldsymbol{\delta}_t|\boldsymbol{z}_t)}{q(\boldsymbol{\delta}_t | \boldsymbol{z}_t, \boldsymbol{u}_t)} \right] \\
    &= \E_{q_{\boldsymbol{z}_t}} \left[\log p(\boldsymbol{o}_t|\boldsymbol{z}_t) +
    \sum_{c=1}^C q(\boldsymbol{\delta}_t=c|\boldsymbol{z}_t) \log p(\boldsymbol{u}_t|\boldsymbol{\delta}_t=c, \boldsymbol{z}_t)\right] - \notag \\
    & KL(q(\boldsymbol{z}_t|\boldsymbol{o}_t), p(\boldsymbol{z}_t)) -  \E_{q_{\boldsymbol{z}_t}} \left[KL(q(\boldsymbol{\delta}_t|\boldsymbol{z}_t, \boldsymbol{u}_t), p(\boldsymbol{\delta}_t|\boldsymbol{z}_t)) \right]
\end{align}}

\section{Training Details}
For the Franka Kitchen task, we keep the hyperparameters the same as the baseline and only change the MDN head to our model.

For the robot writing task, we apply the same hyperparameters to the MDN and our model. Training is done using the ADAM optimizer with a learning rate of 1e-4, a latent state dimension of 128 and the number of skills set to 10. We use an LSTM with all the historical observations to encode the current latent state.

For the fetch push task, we use the same hyperparameters for the MDN and our model. The training is done using the ADAM optimiser with a learning rate of 1e-4, and a latent state dimension of 16. We only use the current latent observations to encode the current latent state.

\section{Results}
\subsection{Franka Kitchen Task}

\subsubsection{Skill Interpretation}
\label{Appdenix: skill interpretation}
Interpretation of skills given task description and learned skill index based on topic modelling.
\begin{table}[!htb]
    \centering
        \begin{tabularx}{\linewidth}{lXX}
        \toprule
        \textbf{Skills} & \textbf{MDN} & \textbf{Our Model}\\
        \midrule
        \textbf{Skill 1} & 'turn', 'rack', 'cheese', 'bottle', 'put' & 'push', 'open', 'drawer', 'plate', 'cabinet' \\
        \textbf{Skill 2} & 'turn', 'stove', 'plate', 'push', 'rack' & 'push', 'open', 'drawer', 'stove', 'plate' \\
        \textbf{Skill 3} & 'turn', 'cheese', 'stove', 'bowl', 'put' & 'turn', 'bowl', 'stove', 'top', 'put' \\
        \textbf{Skill 4} & 'cabinet', 'top', 'bowl', 'open', 'drawer' & 'plate', 'inside', 'bottle', 'open', 'drawer' \\
        \textbf{Skill 5} & 'rack', 'push', 'bottle', 'open', 'drawer' & 'turn', 'stove', 'cheese', 'bowl', 'put' \\
        \bottomrule
        \end{tabularx}
\end{table}


\subsection{Robot Writing Task}
\label{Appendix: robot writing}
\subsubsection{Deployed Trajectories}
The letters labelled by skill index are shown below. The starting point is always $(0, 0)$. We can see the same skill always labels the starting and ending points. This is where the pen lifts or moves down. Additionally, there are also skills like turning and diagonal lifting. The MDN only learns two skills and cannot stop appropriately.

\begin{figure}[!htb]
    \centering
    \subfloat[Our Model: Written letters with segments coloured to indicate the active  skill indices at different points along the trajectory.]{
    \includegraphics[width=0.8\linewidth, trim={10cm 73.5cm 10cm 15cm}, clip]{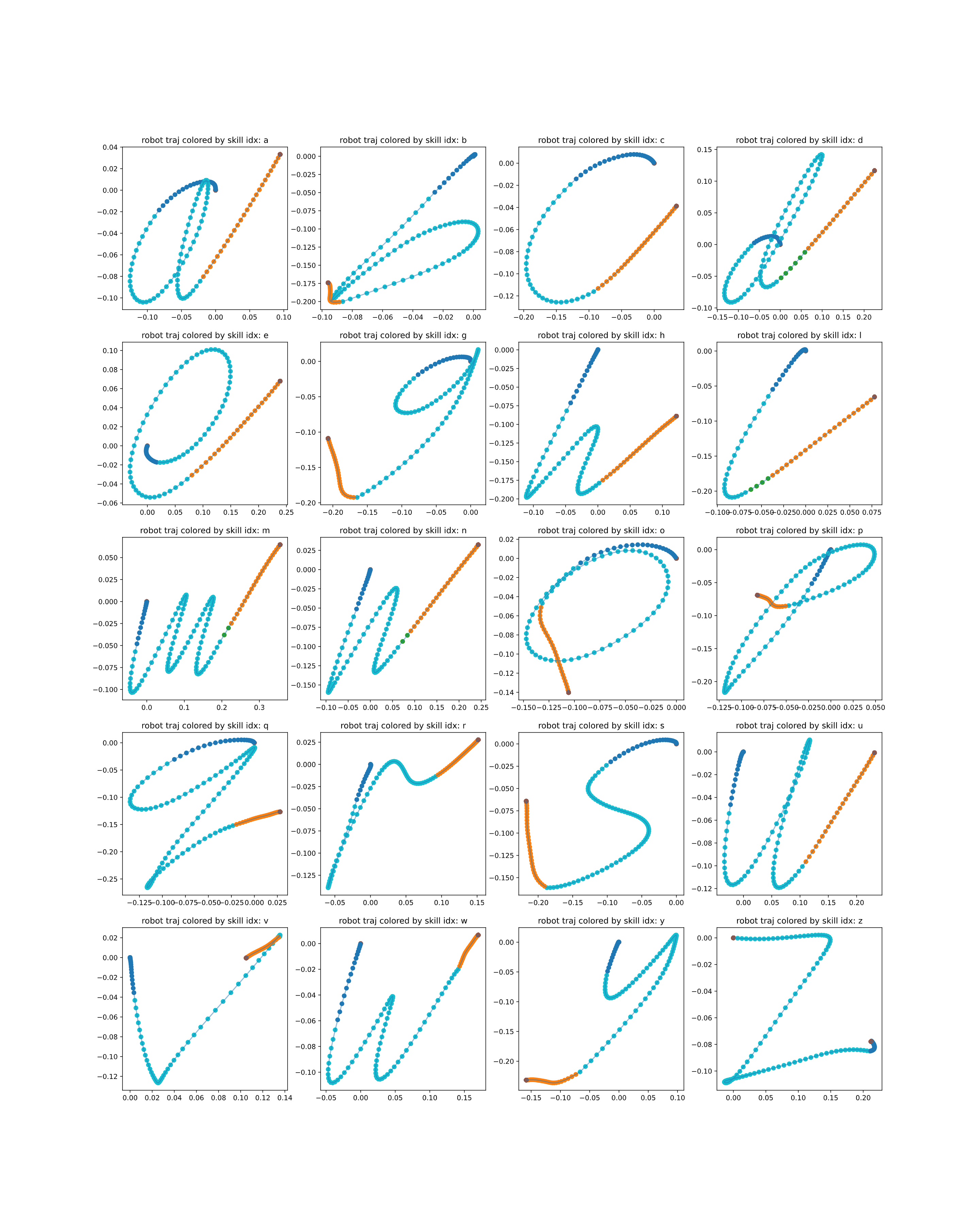}
    }%
    \hfil
    \subfloat[MDN Model: Written letters with segments coloured to indicate the active  skill indices at different points along the trajectory.]{
        \includegraphics[width=0.8\linewidth, trim={10cm 73.5cm 10cm 15cm}, clip]{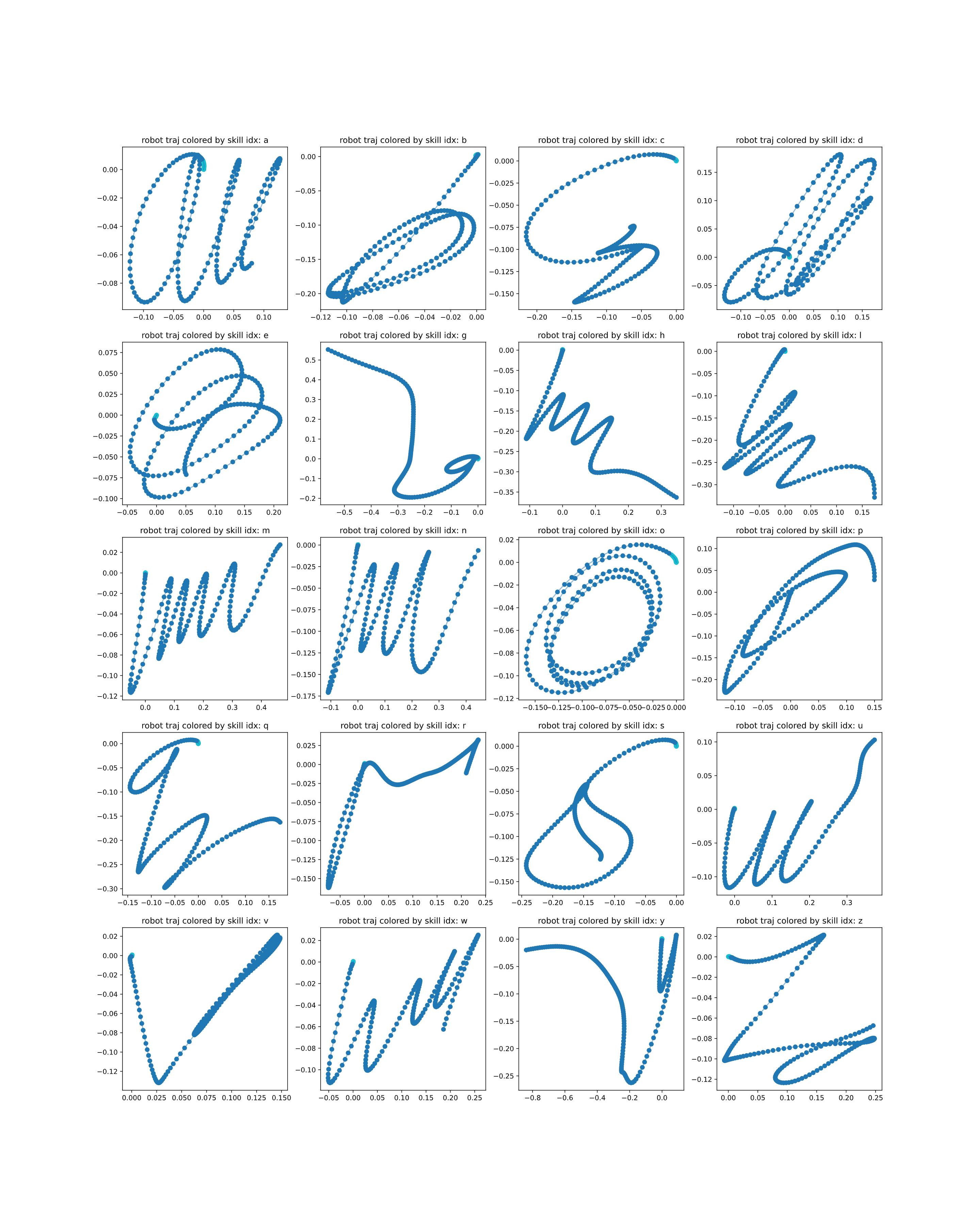}
    }
   \label{fig: letter_by_skill}
\end{figure}

\subsubsection{Velocity and acceleration curve}
Figure 7 presents time-series plots of control signals—both velocity and acceleration—along the x, y, and z axes of the end-effector during the drawing of the letter “a.” The figure compares trajectories generated by the standard MDN model, our enhanced model (MDN + FB + SW), and a human demonstration. Transitions in control signals, corresponding to goal-switching events, occur more frequently in our model (three switches) than in the MDN baseline (one switch). 

The most prominent spikes appear in the z-axis signals, which is expected due to the end-effector establishing,  maintaining and releasing contact with the table surface throughout the writing task. This interaction often results in sharper vertical corrections, particularly during pen-down and pen-lift phases. While our model occasionally introduces higher signal variability, similar patterns are also observed in the human demonstrations. This suggests that our model captures more nuanced, human-like behaviour.

Unlike traditional controllers such as LQR, both the MDN and our model function as trajectory planners, generating waypoint sequences that are later smoothed and executed by low-level controllers. This eliminates the need to hand-craft explicit dynamics models or cost functions, as required in LQR, and enables more flexible, data-driven trajectory generation.

\begin{figure}[!htb]
    \centering
    \includegraphics[width=1\linewidth, trim={0cm 0cm 0cm 0cm}, clip]{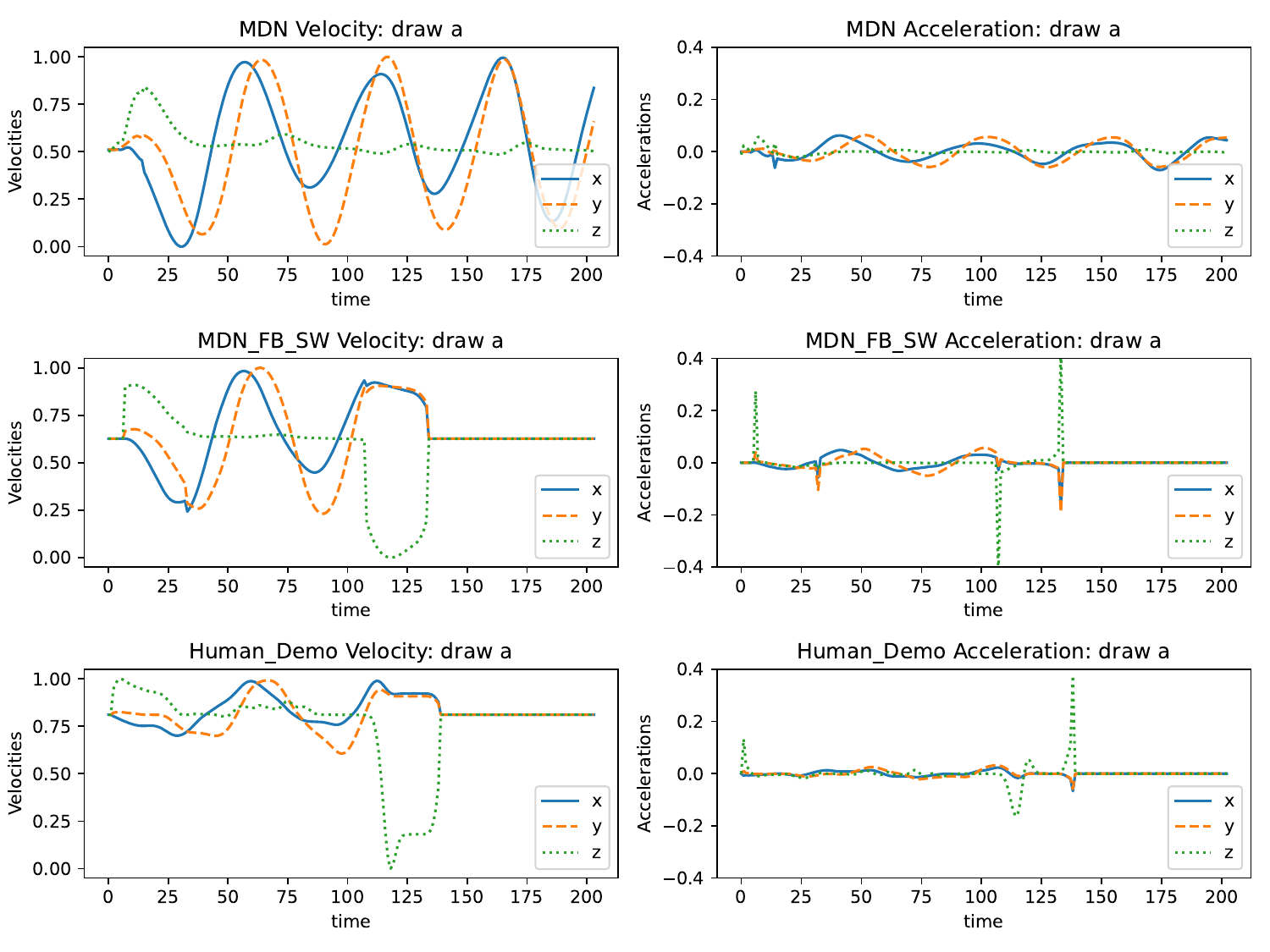}
    \caption{Control signals—velocity and acceleration—in the x, y, and z directions of the end-effector while drawing the letter “a”. The plots compare trajectories generated by the MDN model, our model (MDN + FB + SW), and a human demonstration. Distinct changes in control signals correspond to goal-switching events, with our model exhibiting three switches compared to only one in the MDN model. Our model captures more nuanced, human-like behaviour—even if that sometimes means introducing greater variability, particularly in the z-axis where contact with the table occurs.}
   \label{fig: va_curve_letter_a}
\end{figure}

\bibliography{references,extra_ref}

\newpage

\vfill

\end{document}